\documentclass[10pt,twocolumn,letterpaper]{article}

\usepackage[pagenumbers]{cvpr} 
\usepackage{multirow}
\usepackage{colortbl}
\usepackage{xcolor}
\usepackage{bbm}
\usepackage{makecell}
\usepackage{booktabs}
\usepackage{bbding}

\definecolor{cvprblue}{rgb}{0.21,0.49,0.74}
\usepackage[pagebackref,breaklinks,colorlinks,citecolor=cvprblue]{hyperref}

\def\paperID{4695} 
\def\confName{CVPR}
\def\confYear{2024}

\title{A Unified Framework for Multimodal, Multi-Part Human Motion Synthesis}

\author{Zixiang Zhou\\
{\tt\small zhouzixiang@xiaobing.ai}
\and
Yu Wan\\
{\tt\small wanyu@xiaobing.ai}
\and
Baoyuan Wang\\
{\tt \small wangbaoyuan@xiaobing.ai}
}

\begin{document}

\twocolumn[{
\maketitle
\begin{center}
    \includegraphics[width=1.0\textwidth]{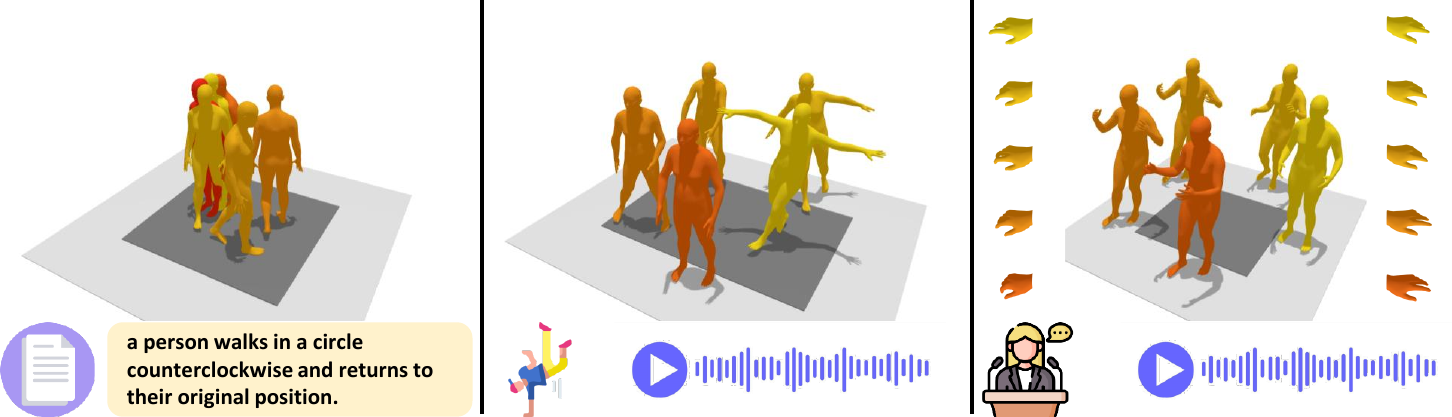}
    \captionof{figure}{Our unified motion synthesis framework supports multimodal and multi-part human motion generation. Left: text-to-motion. Middle: music-to-motion. Right: speech-to-motion.}
    \label{fig:teaser}
\end{center}
}]

\maketitle

\begin{abstract}    
    The field has made significant progress in synthesizing realistic human motion driven by various modalities. Yet, the need for different methods to animate various body parts according to different control signals limits the scalability of these techniques in practical scenarios. In this paper, we introduce a cohesive and scalable approach that consolidates multimodal (text, music, speech) and multi-part (hand, torso) human motion generation. Our methodology unfolds in several steps: We begin by quantizing the motions of diverse body parts into separate codebooks tailored to their respective domains. Next, we harness the robust capabilities of pre-trained models to transcode multimodal signals into a shared latent space. We then translate these signals into discrete motion tokens by iteratively predicting subsequent tokens to form a complete sequence. Finally, we reconstruct the continuous actual motion from this tokenized sequence. Our method frames the multimodal motion generation challenge as a token prediction task, drawing from specialized codebooks based on the modality of the control signal. This approach is inherently scalable, allowing for the easy integration of new modalities. Extensive experiments demonstrated the effectiveness of our design, emphasizing its potential for broad application. Website: \url{https://zixiangzhou916.github.io/UDE-2/}
\end{abstract}

\section{Introduction}

Synthesizing realistic human motion is a pivotal component in many applications, $\textit{e.g.,}$ gaming, virtual reality, and has attracted rising attention in both academia and industry. Recent progress in AI-Generated Content (AIGC) \cite{petrovich2021action,petrovich2022temos,tevet2022human,zhou2023ude,huang2020dance,tseng2023edge,ao2022rhythmic,zhu2023taming} paves a way to generate realistic human motion from various modalities of control signals. These approaches span across handling textual descriptions \cite{petrovich2021action,petrovich2022temos,tevet2022human,zhou2023ude}, music clips \cite{tseng2023edge,siyao2022bailando}, and speech segments \cite{zhu2023taming,ghorbani2023zeroeggs}. Despite the advancements, their primary limitation is the focus on single-modality control signals, overlooking the potential of multi-modal integration.

Among various approaches, UDE\cite{zhou2023ude} can synthesize human motion from text or music using one shared model. While its performance is promising, there are several notable limitations: Firstly, it lacks support for speech as a control signal. Secondly, it focuses solely on the torso, neglecting limb articulation such as hand movements. Thirdly, it consolidates the motions of different domains into one universal codebook — a design strategy we later argue is suboptimal. Finally, updating the motion dataset for one domain adversely impacts the other, compromising efficiency and scalability for further enhancements.

The limitations of current methods become pronounced in practical settings such as gaming or filmmaking, where motions may need to be generated from textual cues, musical scores, or speech. A unified model that can handle these diverse inputs to animate motion would greatly enhance efficiency and practicality. It is equally critical for such a model to prioritize specific body movements depending on the situation; for instance, torso dynamics are crucial in dance sequences, while hand gestures are more prominent during speech. Consequently, a unified approach that can tailor motion synthesis to various body parts, guided by multimodal inputs, is highly valuable.

In this paper, we propose a unified multimodal-driven while-body motion generation framework to overcome the above limitations. 1) We first learn body part-specific and domain-specific VQ-VAEs to quantize motions of different body parts and different domains into corresponding discrete token representations. 2) To interpret the multimodal control signals effectively, we harness the robust representational capabilities of various large-scale pre-trained models, employing them as encoders for the respective signals. These models generate embedding vectors as latent representations, which are then projected to a uniform dimensionality for seamless integration. 3) To learn the mapping from multimodal embedding representations to multi-part body motions, a multimodal fusion transformer is first adopted. It transforms these embeddings into a joint latent space. Subsequently, we frame the motion synthesis as a language generation task, utilizing discrete codebooks for various body parts and domains as different ``vocabulary" in the GPT-like language model\cite{radford2018improving}. Then we predict the next token from the corresponding ``vocabulary"  given conditions and past predictions. 4) We decode the predicted token sequences using corresponding decoders to get the final multi-part body motions. To summarize, our significant contributions are outlined as follows:
\begin{enumerate}
\item We present a pioneering unified human motion generation model that accepts three distinct modalities: text, music, and speech, and enables the synthesis of diverse body part motions, i.e., the torso and left/right hands.

\item We design novel technical components within our model, including a two-stage VQ-VAE tailored for torso motion quantization and a semantic enhancement loss, both of which significantly elevate the effectiveness.

\item We introduce a weight re-initialization for VQ-VAE training and a semantic-aware sampling technique for diverse generations, to get a further performance boost.

\item Comprehensive evaluations of our method across various tasks – text-driven, music-driven, and speech-driven motion generation – demonstrate its competitive performance and robustness in these domains.
\end{enumerate}
These contributions collectively represent a substantial leap forward in the realm of multimodal human motion generation, showcasing the potential for more nuanced and versatile applications.\par

\section{Related Work}

\paragraph{Motion Generation with Text} Recent advances in text-based motion generation have streamlined the process from textual descriptions to motion sequences \cite{petrovich2022temos,guo2022tm2t,tevet2022human,zhou2023ude,zhang2023t2m,lin2023being,petrovich2023tmr,chen2023executing,zhang2023remodiffuse}. The major approaches normally employ VAE-based frameworks to encode text into a latent space from which motions are then reconstructed. Concurrently, diffusion models \cite{tevet2022human,chen2023executing} have been adopted, with conditional variants denoising inputs into motion sequences informed by text. Additionally, the influence of language models is evident, with research by \cite{guo2022tm2t,zhou2023ude,zhang2023t2m,lin2023being} and contemporaries quantizing motions into discrete tokens and utilizing autoregressive techniques for generation, predicting each subsequent motion token based on text and prior tokens.
Moreover, retrieval-based approaches have also been explored \cite{zhang2023remodiffuse,petrovich2023tmr}. Here, a database of motions is queried using text-encoded embeddings to find closely aligned motion sequences in the joint latent space, effectively marrying text inputs with pre-existing motion data.
\vspace{-2mm}

\paragraph{Motion Generation with Music}
The synthesis of dance movements from music has gained significant interest, employing a variety of techniques. Motion graph methods, as in \cite{au2022choreograph} and \cite{chen2021choreomaster} segment dances and build graphs for transition between movements based on dynamic and musical styles. These approaches yield high-fidelity dances but necessitate pre-constructed motion graphs. Alternatively, direct generation techniques like those developed by \cite{li2021ai},\cite{li2022danceformer}, and \cite{tseng2023edge} map music to motion sequences using different architectures: transformer encoder-only, encoder-decoder, and diffusion models, respectively, all focusing on continuous motion prediction from musical inputs. On the other hand, discrete representation is the choice for \cite{siyao2022bailando} and \cite{zhou2023ude}, where dance moves are not directly derived from music. Instead, these models predict discrete token sequences conditioned on the music, based on which regenerates the dance motion.
\vspace{-6mm}

\paragraph{Motion Generation with Speech}In contrast to text- and music-driven synthesis, speech-driven motion synthesis focuses more on gestural rather than torso movements, with considerable research dedicated to addressing this challenge \cite{zhu2023taming,yang2023diffusestylegesture,chemburkar2023discrete,pang2023bodyformer,qi2023emotiongesture,ghorbani2023zeroeggs,alexanderson2023listen,ao2022rhythmic}. Among these methods, diffusion models \cite{zhu2023taming,yang2023diffusestylegesture,alexanderson2023listen,chemburkar2023discrete} stand out for their high diversity in generation, as seen in recent works. Yet, these models often lack the capacity for precise editing or specified generation, such as direct manipulation of latent embeddings. On the other hand, variational generative approaches offer an alternative, as explored by \cite{pang2023bodyformer,qi2023emotiongesture,ghorbani2023zeroeggs,ao2022rhythmic}, where gesture attributes are modulated by distinct latent embeddings produced by dedicated encoders. These models generate gesture movements by integrating all control variables into a continuous representation.
\vspace{-3mm}

\begin{figure*}[h]
    \centering
    \includegraphics[width=\linewidth]{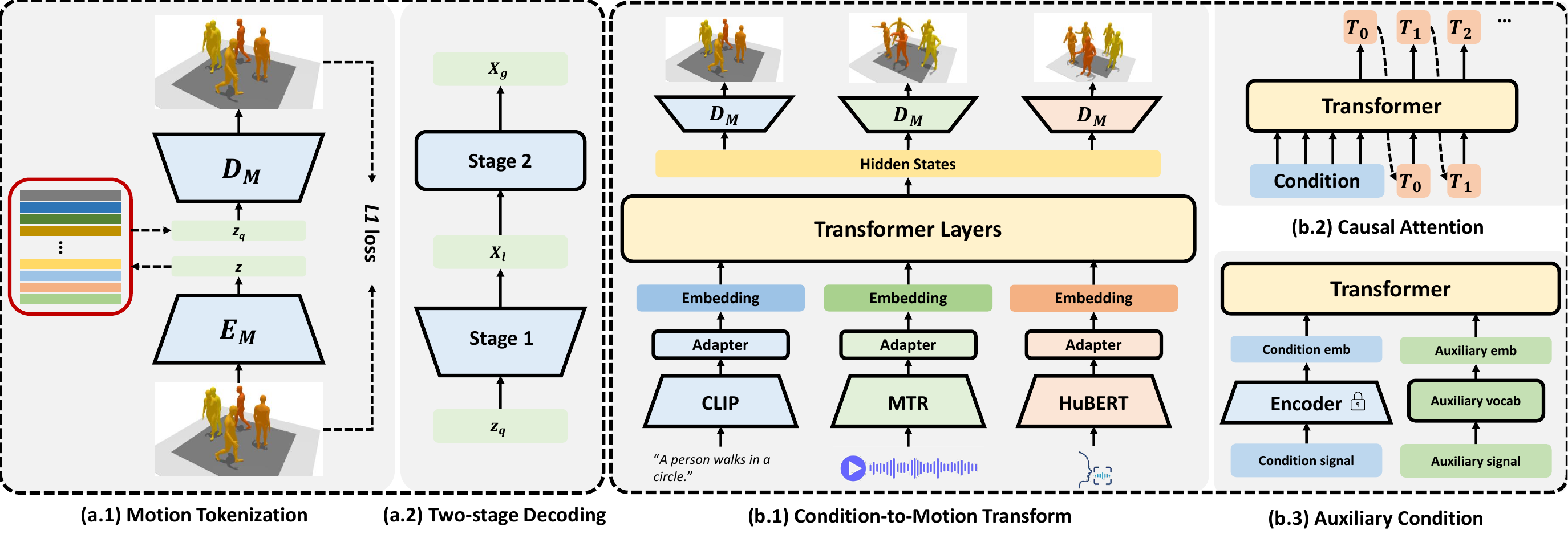}
    \caption{\textbf{Overview.} \textbf{a: Hierarchical Motion Tokenization}, where a two-stage motion decoder is introduced to improve the reconstruction quality of torso movements, compared with conventional VQ-VAE. \textbf{b: Multimodal Multi-party Motion Synthesis}, including a T5\cite{2020t5} -like architecture for condition-to-motion transform(\textbf{b.1}), a causal attention mechanism(\textbf{b.2}) and the fusion of an auxiliary condition (\textbf{b.3}).}
    \label{fig:pipeline-1}
\end{figure*}

\paragraph{The Comparison with Our Work}As can be summarized, current human motion generation techniques tend to specialize in single-modality control signals and fail to address multi-part movements. To generate multi-part body movements conditioned on multimodal control signals, different prior methods have to be engineered, making the framework sophisticated and scalable. In contrast, we introduce a unified framework capable of synthesizing diverse body part motions under the guidance of multimodal inputs.

\section{Method}The proposed method in Fig. \ref{fig:pipeline-1} contains two major modules, 1) a hierarchical motion tokenization module that quantizes whole-body motions to separate codebooks, and 2) a multimodal motion generation module that synthesizes whole-body motion tokens based on multimodal conditions. 

\subsection{Hierarchical Motion Tokenization}We decompose the whole-body pose representation into torso pose and left/right-hand pose representation. Different encoding and decoding strategies are adopted for different body parts.\par
\vspace{-3mm}

\paragraph{Hand Motion Tokenization} Following the design of VQ-VAE in \cite{zhou2023ude}, we learn an encoder $\mathcal{E}(\cdot)$ that transforms hand motion sequence $x \in \mathbb{R}^{T \times c}$ to latent representations $z \in \mathbb{R}^{T' \times d}$, and quantize them by replacing the nearest representation $z^q \in \mathbb{R}^{K \times d}$, where $z^q$ belongs to learned discrete codebook $\mathcal{Z}=\lbrace z_{k}^q | z_{k}^q \in \mathbb{R}^{K \times d} \rbrace$.\par
\vspace{-3mm}

\paragraph{Hierarchical Torso Motion Tokenization}We propose a hierarchical VQ-VAE for torso pose quantization. We use SMPL\cite{SMPL:2015} rotation vector representation for torso poses, each of which is denoted as $x_i=[t_i, r_i]$, where $t_i$ is the root trajectory, and $r_i$ is the rotation vectors. Instead of feeding $x_i=[t_i, r_i]$ to encoder directly, we use a modified representation: $\bar{x}_i=[\bar{t}_i, r_i]$, where $\bar{t}_i$ is calculated as Eq. \ref{eq:trajectory}:

\begin{equation}
    \label{eq:trajectory}
    \bar{t}_{i}=\left\{
    \begin{aligned}
    \relax[0, 0, 0] & , & i = 0, \\
    t_i - t_{i-1} & , & i > 0.
    \end{aligned}
    \right.
\end{equation}

This modification focuses on pose and local trajectory, making the learned discrete representation more robust to global trajectory shifting. We express the encoding as: $z^q = \mathcal{E}(\bar{x})$, where $z \in \mathbb{R}^{T' \times d}$ is the encoded latent representation. We follow the quantization process in \cite{zhou2023ude} to obtain the quantized representation $z^q \in \mathbb{R}^{K \times d}$.\par

During decoding, we first reconstruct local poses $\bar{y} \in \mathbb{R}^{T \times c}$ as: $\bar{y} = \mathcal{D}_l(z^q)$. For each pose $\bar{y}_i$, it is expressed as: $\bar{y}_i = [\bar{t}_i, r_i]$. Since we can estimate $t_i$ from $\bar{t}_i$ according to Eq.\ref{eq:trajectory}, we can, hence, obtain sub-optimal global pose as $\tilde{y} = \mathcal{R}(\bar{y})$, where $\mathcal{R}(\cdot)$ is the reverse process of Eq.\ref{eq:trajectory}. We refine $\tilde{y}$ using a 1D U-net network to obtain the final pose: $y = \mathcal{D}_g(\tilde{y})$. The two-stage decoding could be expressed as: $y = \mathcal{D}_g(\mathcal{R}(\mathcal{D}_l(z_q)))$. The loss of two-stage VQ-VAE is calculated as Eq. \ref{eq:vq-loss}:
\vspace{-3mm}

\begin{equation}
    \mathcal{L}_{VQ} = \mathcal{L}_{rec} + \beta_1 \| sg[z] - z_q \| + \beta_2 \| z - sg[z^q] \|
    \label{eq:vq-loss}
\end{equation}

where $\mathcal{L}_{rec} = \alpha_1\mathcal{L}_{rec}^l + \alpha_2\mathcal{L}_{rec}^{\tilde{g}} + \alpha_3\mathcal{L}_{rec}^{g}$ is the reconstruction term that contains three sub-items, 1) $\mathcal{L}_{rec}^l = \| \bar{x} - \bar{y} \|$ minimizes the distance between local poses and reconstructed local poses, 2) $\mathcal{L}_{rec}^{\tilde{g}} = \| x - \tilde{y} \|$ minimizes the distance between global poses and reconstructed sub-optimal global poses, and 3) $\mathcal{L}_{rec}^{g} = \| x - y \|$ minimizes the global poses and reconstructed global poses. The definition of other two terms $\| sg[z] - z_q \|$, $\| z - sg[z^q] \|$ follow \cite{zhou2023ude}.\par
\vspace{-3mm}

\paragraph{Embedding Weights Re-initialization}We define the terminology \textit{``activated"} as if the weight of token embedding is updated during optimization, and it could be decoded back to a valid motion segment. It is non-trivial to tune the factor $\beta_1$, $\beta_2$ in Eq. \ref{eq:vq-loss} to increase the activation rate. We propose a novel training technique, called \textit{"weight re-initialization"} to solve this problem. For every $k$ step during training, we count the activation rate of every token in the codebook and sort them in descending order, the sorted tokens embeddings are: $[e_i^1, e_j^2, ..., e_k^n]$, where subscripts $i,j,k$ indicate their original order, and superscripts indicate their sorted order. And the activation rate after sorting follows: $r(e^1) > r(e^2) > ... > r(e^n)$. We re-initialize the weight of $e^k$ as $e_k\leftarrow e^{n-k}+\delta$ if $r(e^k) < \tau$, where $\delta \sim \mathcal{N}(0, \sigma{I})$, and $\sigma$ is a constant with very small value.\par

\begin{figure}[h]
    \centering
    \includegraphics[width=1.0\linewidth]{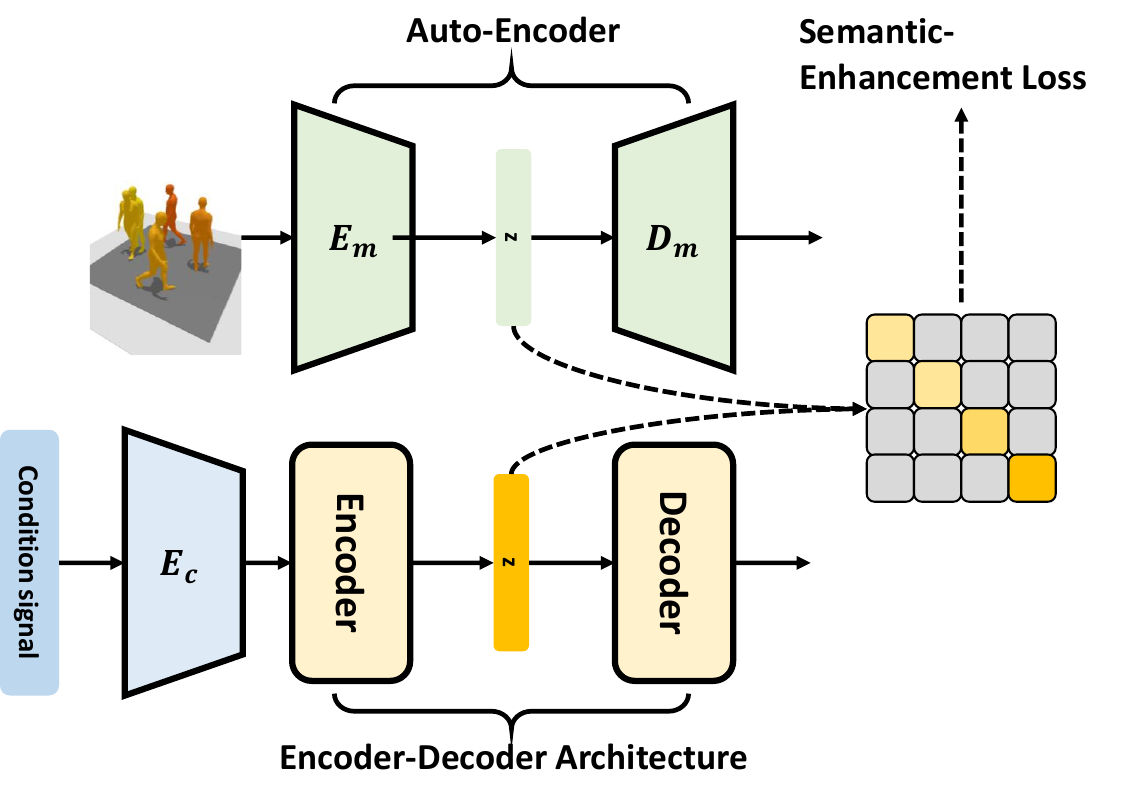}
    \caption{\textbf{Pipeline of semantic enhancement.} We propose a novel method to enhance the degree of semantic distinction of condition embedding. We pre-train a motion auto-encoder as prior and use it to enhance the degree of semantic distinction of the hidden states of the encoder part.}
    \label{fig:semantic-enhance}
\end{figure}
\vspace{-3mm}

\begin{figure*}[h]
    \centering
    \includegraphics[width=\linewidth]{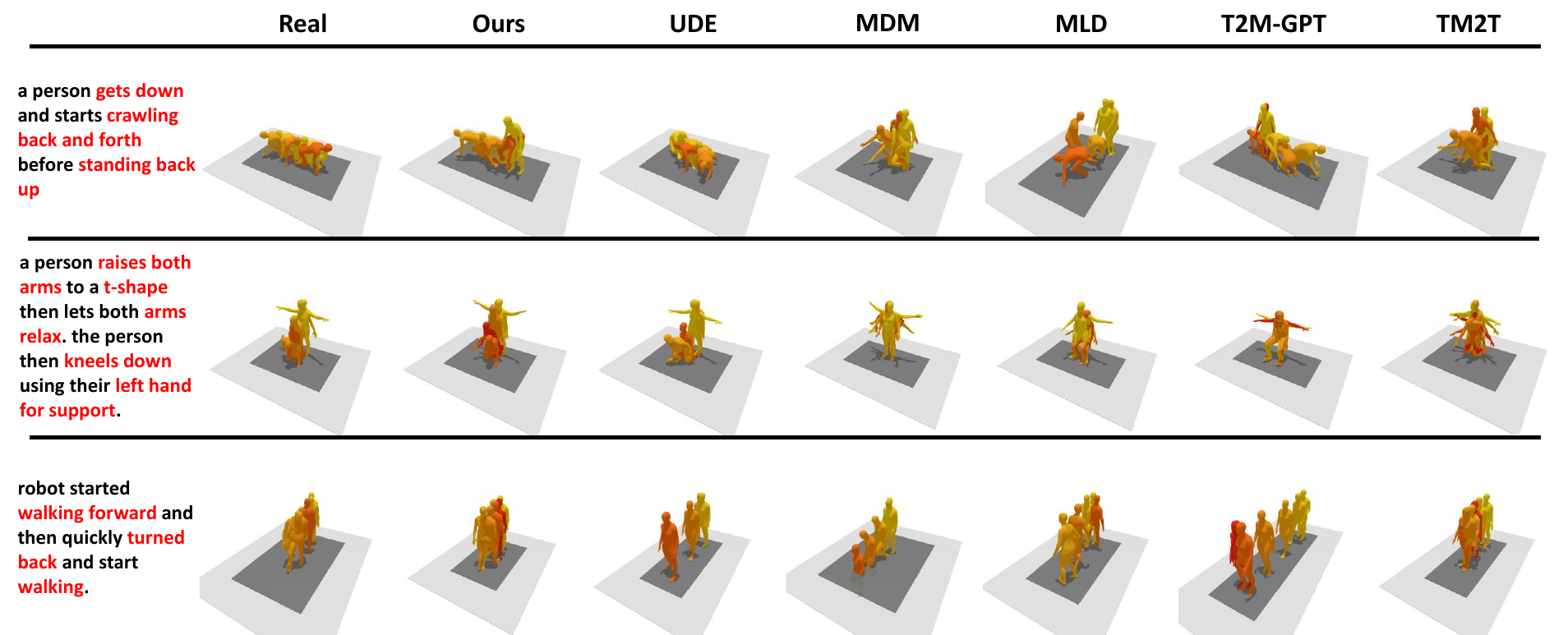}
    \caption{\textbf{Visual Comparison on Text-to-Motion Task.} We evenly show 5 poses for each motion to compare ours with SOTAs on text-to-motion tasks. The condition language descriptions are shown in the very left column. Keywords are highlighted.}
    \label{fig:result-t2m}    
\end{figure*}

\begin{table*}[]
    \centering
    \resizebox{0.8\linewidth}{!}{
        \begin{tabular}{c|ccccc|ccc}
            \hline
            \multirow{2}{*}{Method} & \multicolumn{5}{c|}{R Precision} & \multirow{2}{*}{FID $\downarrow$} & \multirow{2}{*}{Div $\rightarrow$} & \multirow{2}{*}{MM $\uparrow$}  \\
            \cline{2-6}
             & Top-1 $\uparrow$ & Top-2 $\uparrow$ & Top-3 $\uparrow$ & Top-4 $\uparrow$ & Top-5 $\uparrow$ \\
            \hline
            GT & 63.49 & 78.50 & 84.66 & 88.14 & 90.35 & 0.082 & 14.45 & - \\
            \hline
            TM2T\cite{guo2022tm2t} & $42.27^{\pm{1.137}}$ & $57.04^{\pm{1.200}}$ & $65.10^{\pm{1.293}}$ & $70.58^{\pm{1.285}}$ & $74.82^{\pm{1.313}}$ & $10.80^{\pm{0.065}}$ & $13.77^{\pm{0.058}}$ & $1.62^{\pm{0.342}}$ \\
            MDM\cite{tevet2022human} & $48.88^{\pm{1.121}}$ & $65.53^{\pm{1.200}}$ & $72.81^{\pm{1.293}}$ & $78.12^{\pm{1.335}}$ & $81.47^{\pm{1.281}}$ & $4.53^{\pm{0.034}}$ & $13.88^{\pm{0.034}}$ & \cellcolor{yellow!15}$7.12^{\pm{0.071}}$ \\
            T2M-GPT\cite{zhang2023t2m} & \cellcolor{yellow!15}$51.36^{\pm{1.187}}$ & \cellcolor{yellow!15}$64.85^{\pm{1.217}}$ & \cellcolor{red!15}$75.88^{\pm{1.140}}$ & \cellcolor{red!15}$78.75^{\pm{1.136}}$ & \cellcolor{red!15}$82.32^{\pm{1.177}}$ & $6.67^{\pm{0.061}}$ & $13.65^{\pm{0.219}}$ & $5.65^{\pm{0.229}}$ \\
            MLD\cite{chen2023executing} & $48.85^{\pm{0.996}}$ & $65.52^{\pm{1.096}}$ & $73.21^{\pm{1.326}}$ & $78.52^{\pm{1.143}}$ & $80.97^{\pm{1.246}}$ & $11.85^{\pm{0.097}}$ & $13.64^{\pm{0.219}}$ & $4.63^{\pm{1.655}}$ \\ 
            UDE\cite{zhou2023ude} & $46.69^{\pm1.152}$ & $61.69^{\pm{1.259}}$ & $69.82^{\pm{1.239}}$ & $75.04^{\pm{1.240}}$ & $78.61^{\pm{1.236}}$ & \cellcolor{yellow!15}$2.57^{\pm{0.019}}$ & \cellcolor{red!15}$14.25^{\pm{0.034}}$ & $6.83^{\pm{0.121}}$ \\ 
            \hline
            Ours & \cellcolor{red!15}$54.48^{\pm{0.340}}$ & \cellcolor{red!15}$68.02^{\pm{0.280}}$ & \cellcolor{yellow!15}$74.50^{\pm{0.280}}$ & \cellcolor{yellow!15}$78.62^{\pm{0.240}}$ & \cellcolor{yellow!15}$81.63^{\pm{0.230}}$ & \cellcolor{red!15}$2.54^{\pm{0.045}}$ & \cellcolor{yellow!15}$14.72^{\pm{0.034}}$ & \cellcolor{red!15}$8.48 ^{\pm{0.297}}$ \\
            \hline
        \end{tabular}
    }
    \caption{\textbf{Results of text-to-motion.} We compare the results of text-to-motion generation between ours and the SOTA methods. Our method achieves better semantic relevance, fidelity, and diversity performances. $\colorbox{red!15}{\rm Indicate best results}, \colorbox{yellow!15}{\rm indicates second best results}$.}
    \label{tab:main-t2m}
\end{table*}

\begin{figure*}[h]
    \centering
    \includegraphics[width=0.8\linewidth]{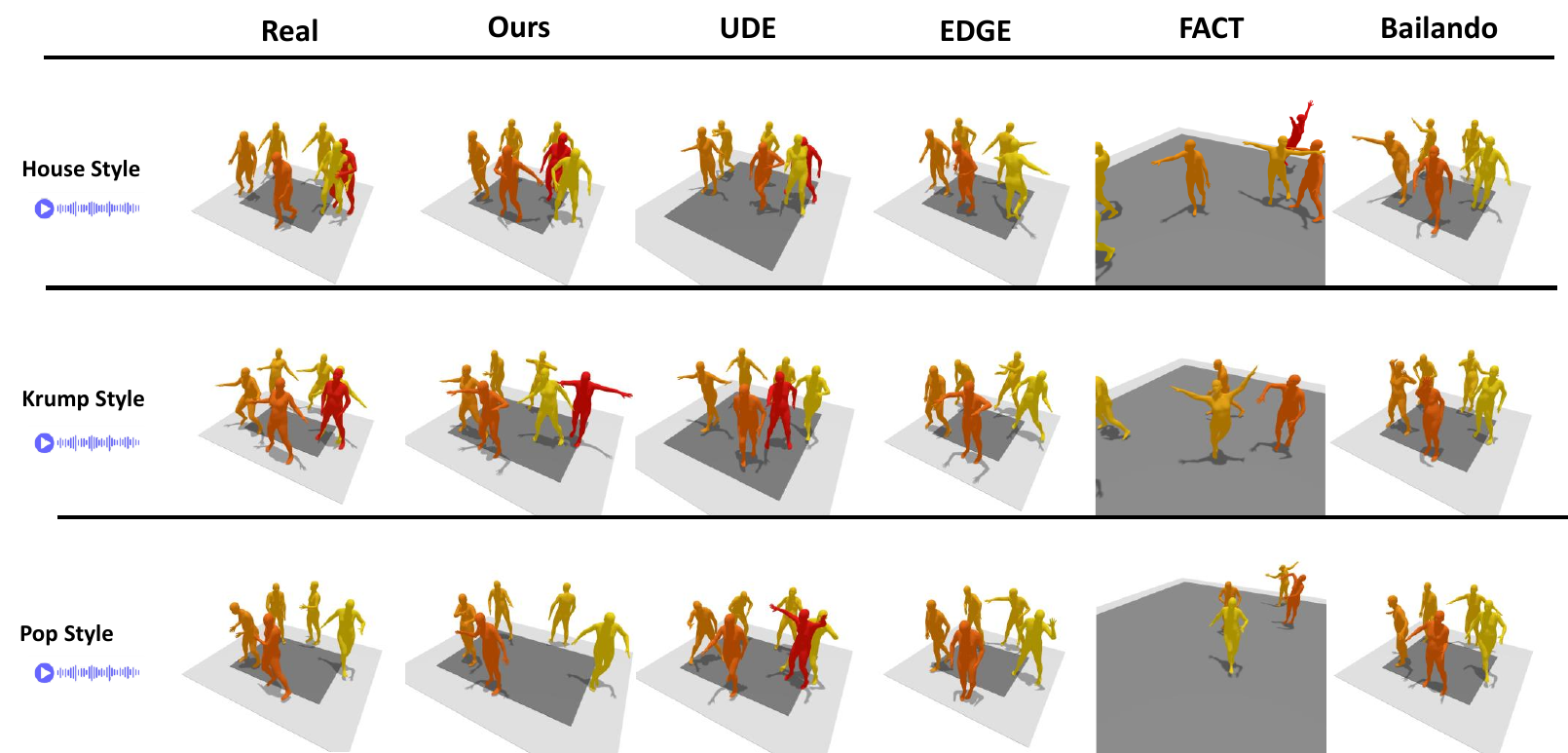}
    \caption{\textbf{Visual Comparison on Music-to-Motion Task.} To ease the visualization, we modify the trajectory of each dance movement and present 5 poses per sequence.}
    \label{fig:result-a2m}    
\end{figure*}

\begin{table*}
    \centering
    \resizebox{0.8\linewidth}{!}{
        \begin{tabular}{c|cccc|cccc}
            \hline
            \multirow{2}{*}{Method} & \multicolumn{4}{c|}{MDSC} & \multirow{2}{*}{FID $\downarrow$} & \multirow{2}{*}{Div $\rightarrow$} & \multirow{2}{*}{MM $\uparrow$} & \multirow{2}{*}{BAS $\uparrow$} \\
            \cline{2-5}
             & Acc. $\uparrow$ & Top-1 Retr. $\uparrow$ & I2I. $\downarrow$ & Simi. $\uparrow$ & & & & \\ 
            \hline
            GT & 38.32 & 37.65 & 0.72 & 0.48 & 1.03 & 2.19 & - & 0.237 \\ 
            \hline
            FACT\cite{li2021ai} & 17.67 & 17.96 & 0.92 & 0.01 & 2.43 & 2.89 & - & 0.221 \\ 
            Bailando\cite{siyao2022bailando} & \cellcolor{yellow!15}37.04 & \cellcolor{yellow!15}38.01 & \cellcolor{yellow!15}0.74 & 0.32 & 1.04 & \cellcolor{red!15}{2.12} & - & \cellcolor{yellow!15}0.233 \\
            UDE\cite{zhou2023ude} & $20.69^{\pm{4.231}}$ & $19.84^{\pm{3.902}}$ & 0.89 & 0.17 & \cellcolor{red!15}$0.38^{\pm{0.015}}$ & $2.04^{\pm{0.014}}$ & \cellcolor{red!15}$1.85^{0.007}$ & $0.231^{\pm{0.019}}$ \\
            EDGE\cite{tseng2023edge} & $23.93^{\pm{2.404}}$ & $23.69^{\pm{2.211}}$ & 0.86 & \cellcolor{yellow!15}0.41 & $1.17^{\pm{0.015}}$ & $1.85^{\pm{0.016}}$ & $1.68^{\pm{0.092}}$ & \cellcolor{red!15}{0.260} \\ 
            \hline
            Ours & \cellcolor{red!15}$54.25^{\pm{4.564}}$ & \cellcolor{red!15}$53.89^{\pm{4.119}}$ & \cellcolor{red!15}0.56 & \cellcolor{red!15}0.60 & \cellcolor{yellow!15}$0.46^{\pm{0.023}}$ & \cellcolor{yellow!15}$2.06^{\pm{0.021}}$ & \cellcolor{yellow!15}$1.75^{\pm{0.028}}$ & $0.222^{\pm{0.025}}$ \\
            \hline            
        \end{tabular}
    }
    \caption{\textbf{Results of music-to-motion.} We compare the style consistency, motion quality, and diversity between ours and previous methods and the results suggest that our method achieves the highest consistency. $\colorbox{red!15}{\rm Indicate best results}, \colorbox{yellow!15}{\rm indicates second best results}$.}
    \label{tab:main-a2m}
\end{table*}

\begin{table}
    \centering
    \resizebox{1.0\linewidth}{!}{
        \begin{tabular}{c|cc|ccc}
            \hline
            \multirow{2}{*}{Method} & \multicolumn{2}{c|}{ID Consis.} & \multirow{2}{*}{FID $\downarrow$} & \multirow{2}{*}{Div $\rightarrow$} & \multirow{2}{*}{MM $\uparrow$} \\
            \cline{2-3}
            & Acc. $\uparrow$ & I2I. $\downarrow$ & & & \\ 
            \hline
            GT & 99.97 & 0.36 & 0.36 & 2.26 & - \\ 
            \hline
            DSG\cite{yang2023diffusestylegesture} & \cellcolor{red!15}$99.91^{\pm{0.000}}$ & \cellcolor{red!15}$0.44^{\pm{0.000}}$ & \cellcolor{red!15}$0.55^{\pm{0.000}}$ & \cellcolor{yellow!15}$2.18^{\pm{0.000}}$ & \cellcolor{yellow!15}$1.80^{\pm{0.005}}$ \\ 
            \hline
            Ours & \cellcolor{yellow!15}$93.16^{\pm{7.900}}$ & \cellcolor{yellow!15}$0.46^{\pm{0.051}}$ & \cellcolor{yellow!15}$0.75^{\pm{0.054}}$ & \cellcolor{red!15}$2.37^{\pm{0.018}}$ & \cellcolor{red!15}$2.01^{\pm{0.006}}$ \\ 
            \hline
        \end{tabular}
    }
    \caption{\textbf{Results of speech-to-motion.} We evaluate the ID consistency, motion quality, and diversity between our and SOTA methods. $\colorbox{red!15}{\rm Indicate best results}, \colorbox{yellow!15}{\rm indicates second best results}$.}
    \label{tab:main-s2m}
\end{table}

\begin{figure}
    \centering
    \includegraphics[width=1.0\linewidth]{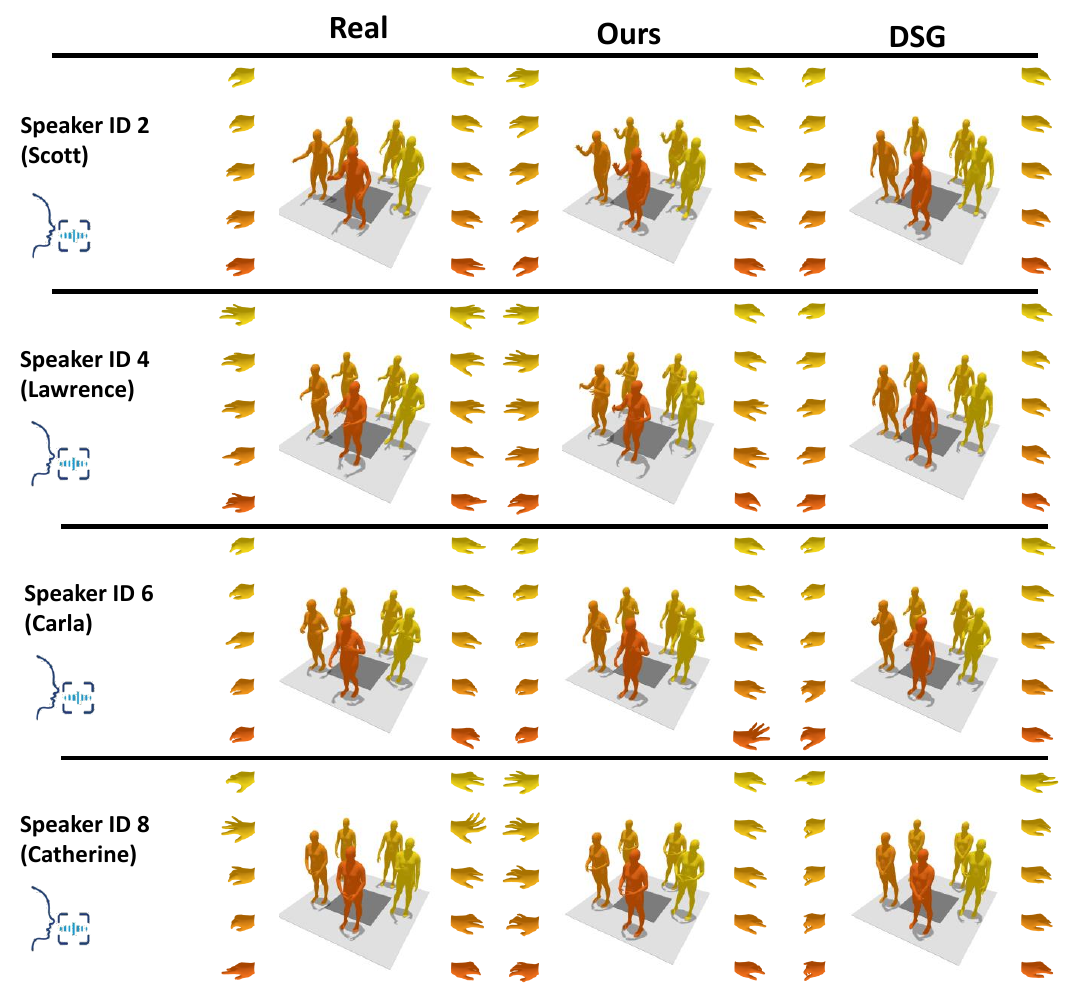}
    \caption{\textbf{Visual Comparison on Speech-to-Motion Task.} To ease the visualization, we modify the trajectory of each dance movement and present 5 poses per sequence. We enlarge the display of hand motions and show them on both sides of the whole-body motion display.}
    \label{fig:result-s2m}
\end{figure}


\subsection{Multimodal Multi-Part Motion Generation}As illustrated in Fig. \ref{fig:pipeline-1} (b.1), our method supports generating whole-body(torso, left/right hand) motion with three different input modalities. We describe the details below.\par
\vspace{-3mm}

\paragraph{Multimodal Alignment}We leverage the power of large-scale pre-trained models as modality encoders. We use CLIP text encoder\cite{radford2021learning} to understand and encode language descriptions. We adopt MTR\cite{doh2023toward} to encode the music pieces, it is a multimodal model pre-trained on text-music pair data with contrastive loss. We employ HuBERT\cite{hsu2021hubert}, a self-supervised trained model to learn speech embeddings. For all encoders, we adopt the hidden states of their last layers as the latent embedding sequence. We denote, for modality $p$, the input condition is $x_p \in \mathbb{R}^{T_p \times c_p}$, the latent embedding is $e_p \in \mathbb{R}^{T'_p \times c'_p}$. We introduce adapter layers to align multimodal embeddings as: $\mathcal{A}_p(e_p): \mathbb{R}^{T'_p \times c'_p} \rightarrow \mathbb{R}^{T'_p \times d}$. The multimodal alignment could be expressed as: $e_p = \mathcal{A}_p(\mathcal{E}_m(x_p))$.\par
\vspace{-3mm}

\paragraph{Hierarchical Condition to Motion Tokens Transform}We adopt an encoder-decoder transformer architecture to transform multimodal conditions to motion tokens. The encoder part takes as input the aligned embeddings $e_p \in \mathbb{R}^{T'_p \times d}$. We pad special tokens to unify the lengths. We extract context information with full attention mechanism and transform it to latent embedding $\tilde{e}_p \in \mathbb{R}^{T'_p \times d}$. Then we employ a transformer decoder to map $\tilde{e}_p$ to motion tokens. We apply causal masks to prevent future information from being leaked in training. We predict the next token successively when generating. We propose a hierarchical decoder architecture to generate tokens of different body parts. Assuming that torso movements depend on the condition and past torso movements, the transform of torso motion tokens is $p_{\theta_t}(x_t) = \prod_{i=1}^{n}p_{\theta_t}(x_{t,i}|x_{t,<i},C)$, where $C$ is condition. Same assumption holds for hands movements as: $p_{\theta_h}(x_h) = \prod_{i=1}^{n}p_{\theta_h}(x_{h,i}|X_{h,<i},C)$. In our hierarchical design, we adopt a base decoder to learn the common latent information. We use separate head decoders to learn part-specific and modality-specific motion tokens. Therefore, the hierarchical condition-to-motion transform could be written as:
\vspace{-5mm}

\begin{equation}
    p(x_{b,p}) = \prod_{i=1}^np_{\theta}(x_{b,p,i}|X_{b,p,<i},C_p)p_{\theta_{b,p}}(x_{b,p,i}|X_{b,p,<i},C_p)
    \label{eq:auto-regressive-hierarchical}
\end{equation}

In Eq. \ref{eq:auto-regressive-hierarchical}, $b$ denotes body parts, and $p$ denotes modality of the condition, $\theta$ denotes the parameters of base decoder, and $\theta_{b,p}$ denotes the parameters of head decoders.\par
\vspace{-3mm}

\paragraph{Separate Motion Vocabulary}We use separate vocabulary for different body parts subject to different modalities. The use of separate vocabulary brings several advantages, 1) it avoids sampling out-of-domain motion tokens, 2) updating the vocabulary of one domain will not affect the others, and 3) it is efficient to introduce additional modality to our pipeline as a plug-in.\par
\vspace{-3mm}

\paragraph{Alignment of Auxiliary Condition}Our method supports taking any auxiliary conditions as a supplement. For instance, when generating gesture motion from speech, the persona of characters matters. We introduce trainable auxiliary vocabulary to the encoder part as $\mathcal{Z}_{aux} = \lbrace e_{aux}|e_{aux} \in \mathbb{R}^{K \times d} \rbrace$, where $K$ is the number of auxiliary conditions, and $d$ is the latent dimension. We append the auxiliary embedding $e_{aux}$ to dominant condition embeddings $e_m$ as: $e_{fused}=[e_m, e_{aux}]$, shown in Fig. \ref{fig:pipeline-1} (b.3).\par
\vspace{-3mm}

\paragraph{Semantic Enhancement}Semantic meaning of condition signal plays a vital role in generating human motion. It is self-explanatory that conditions with opposite semantic meanings should correspond to human movements of obvious difference. For instance, the human motions described by $C_1$: ``\textit{a person is running forward}" and $C_2$:``\textit{a person is running backward}" should present remarkable difference, while the motion described by $C_3$:``\textit{a person is walking forward}" is expected to be more similar with the one described by $C_1$ than $C_2$. We denote the motion generated by condition $C$ as $x_c$, the similarity between two motion sequences and conditions are $s_m = \langle x_{c_1}, x_{c_2}\rangle$ and $s_c = \langle c_1, c_2\rangle$. We empirically found that for any two pairs of $\lbrace x_c, c\rbrace$, $s_m \propto s_c$ does not hold. It is particularly obvious for text-to-motion scenarios. We propose a novel semantic enhancement loss to facilitate the alignment between motion and condition. As shown in Fig. \ref{fig:semantic-enhance}, we pre-train a motion auto-encoder and use its encoder to encode motion as $e_m \in \mathbb{R}^{1 \times d}$. We denote $e_c\in\mathbb{R}^{1\times d}$ as temporal average of condition embeddings $\tilde{e}_p\in\mathbb{R}^{T'_p \times d}$ obtained by condition-to-motion encoder transformer. We align $e_m$ and $e_c$ in latent space by maximizing the cross-modality similarity. Specifically, we fix the motion encoder and optimize the condition-to-motion encoder to align $e_c$ to $e_m$. This benefits learning motion-aware representation from the condition. Our semantic enhancement loss is:
\vspace{-3mm}

\begin{equation}
    \mathcal{L}_{sem} = \frac{1}{N}\Sigma_{i=1}^N(1 - \langle s_m^i, s_c^i \rangle)
    \label{eq:semantic-enhance}
\end{equation}

\paragraph{Semantic-Aware Sampling}We propose a simple yet efficient sampling approach called \textit{``semantic-aware sampling"} to boost the semantic correlation while maintaining generation diversity. We found that the token embedding similarity is proportional to motion dynamic similarity. Instead of sampling a token from a probability distribution, we sample one token using a different approach. Suppose token $y_i$ corresponds to the largest probability $p(x_i|x_1,x_2,...,x_i-1,C)$, and its embedding in codebook is $z^q_i$. We calculate the pairwise distance between $z_i^q$ and $\{z_j^q\}_{j=1}^K$ as: $d_{ij}(z_i^q, z_j^q) = \|z_i^q-z_j^q\|_2$. Then convert the distance to weight factors as: 
\vspace{-3mm}

\begin{equation}
    w_{ij} = \frac{exp(-d_{ij}/t)}{\Sigma_{j=1}^K{exp(-d_{ij}/t}})
    \label{eq:sas-weight}
\end{equation}
\vspace{-3mm}

where $t$ is the temperature coefficient. Multiplying weight $w$ by probability $p$ adjusts the distribution by reducing the probability that dynamically irrelevant tokens have been sampled. Hence, increasing the semantic consistency.\par
\section{Experiments}

\begin{figure*}
    \centering
    \includegraphics[width=0.9\linewidth]{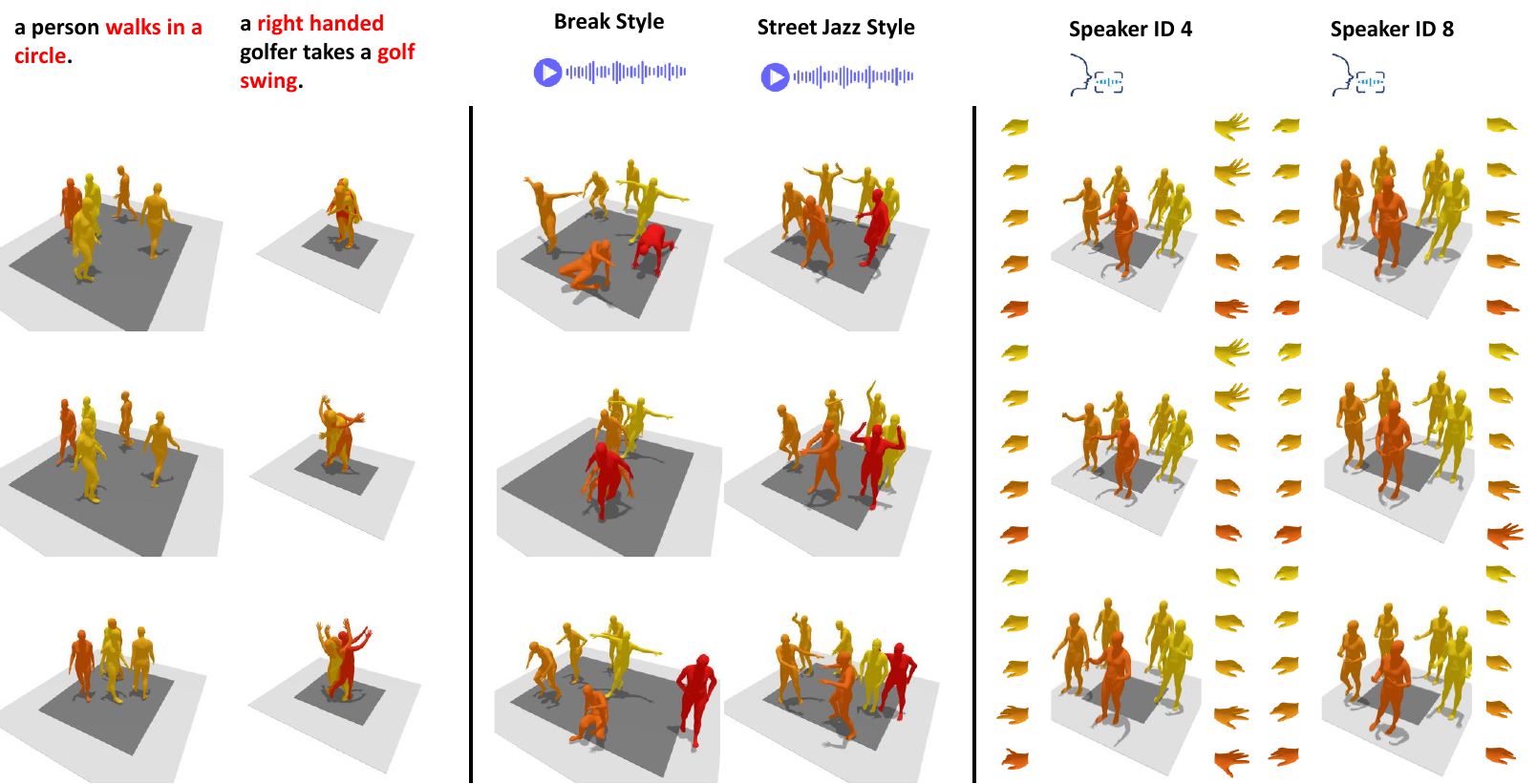}
    \caption{\textbf{Diverse Generated Motions.} We generate 3 motion sequences from the same condition to present that our method has promising ability of diverse generation. \textbf{Left}: text-to-motion. \textbf{Middle}: music-to-motion. \textbf{Right}: speech-to-motion.}
    \label{fig:result-diversity}
\end{figure*}

\begin{table*}
    \centering
    \resizebox{0.9\linewidth}{!}{
        \begin{tabular}{cc|cccc|cccc|cccc}
            \hline
            \multicolumn{2}{c|}{Method} & \multicolumn{4}{c|}{HumanML3D\cite{guo2022tm2t}} & \multicolumn{4}{c|}{AIST++\cite{li2021ai}} & \multicolumn{4}{c}{BEAT\cite{liu2022beat}} \\
            \hline
             Two-stage & Re-init. & FID $\downarrow$ & MPJPE $\downarrow$ & MPJVE $\downarrow$ & MPJAE $\downarrow$ & FID $\downarrow$ & MPJPE $\downarrow$ & MPJVE $\downarrow$ & MPJAE $\downarrow$ & FID $\downarrow$ & MPJPE $\downarrow$ & MPJVE $\downarrow$ & MPJAE $\downarrow$ \\
             \hline
             \checkmark & \XSolidBrush & 1.807 & 0.100 & 0.012 & 0.012 & 5.449 & 0.154 & 0.017 & 0.013 & 0.682 & \cellcolor{red!15}0.018 & 0.0018 & 0.0013 \\
             \XSolidBrush & \checkmark & 1.693 & 0.072 & 0.010 & 0.008 & 5.278 & 0.090 & 0.014 & 0.010 & 0.726 & 0.019 & 0.0018 & 0.0012 \\
             \checkmark & \checkmark & \cellcolor{red!15}1.167 & \cellcolor{red!15}0.054 & \cellcolor{red!15}0.007 & \cellcolor{red!15}0.004 & \cellcolor{red!15}5.193 & \cellcolor{red!15}0.087 & \cellcolor{red!15}0.012 & \cellcolor{red!15}0.007 & \cellcolor{red!15}0.518 & \cellcolor{red!15}0.018 & \cellcolor{red!15}0.0017 & \cellcolor{red!15}0.0011 \\
             \hline        
        \end{tabular}
    }
    \caption{\textbf{Ablation Study on Variants VQ-VAE Designs.} The results suggest two-stage design and re-initialization strategy improve the performance well. $\colorbox{red!15}{\rm Indicate best results}$.}
    \label{tab:ablation-vq}
\end{table*}

\begin{table*}
    \centering
    \resizebox{0.85\linewidth}{!}{
        \begin{tabular}{c|cccc|cccc|cccc}
            \hline
            \multirow{2}{*}{Method} & \multicolumn{4}{c|}{Text-to-Motion} & \multicolumn{4}{c|}{Music-to-Motion} & \multicolumn{4}{c}{Speech-to-Motion} \\
            \cline{2-13}
            & Top-1 $\uparrow$ & Top-3 $\uparrow$ & FID $\downarrow$ & MM $\uparrow$ & Acc. $\uparrow$ & I2I. $\downarrow$ & FID $\downarrow$ & MM $\uparrow$ & Acc. $\uparrow$ & I2I. $\downarrow$ & FID $\downarrow$ &MM $\uparrow$ \\
            \hline
            w/o SemE. & 53.99 & 74.22 & 2.57 & - & 54.14 & 0.56 & \cellcolor{red!15}0.44 & - & \cellcolor{red!15}94.48 & 0.50 & \cellcolor{red!15}0.66 & - \\
            w/ SemE. & \cellcolor{red!15}55.06 & \cellcolor{red!15}75.60 & \cellcolor{red!15}2.56 & - & \cellcolor{red!15}54.29 & \cellcolor{red!15}0.55 & 0.60 & - & 93.92 & \cellcolor{red!15}0.49 & 0.83 & - \\
            \hline
            w/o SaS. & 52.71 & 73.06 & \cellcolor{red!15}2.48 & \cellcolor{red!15}9.02 & 53.85 & 0.57 & 0.53 & \cellcolor{red!15}1.86 & 92.61 & 0.49 & 0.89 & \cellcolor{red!15}2.02 \\
            w/ Sas. & \cellcolor{red!15}54.48 & \cellcolor{red!15}74.50 & 2.54 & 8.48 & \cellcolor{red!15}54.25 & \cellcolor{red!15}0.56 & \cellcolor{red!15}0.46  & 1.75 & \cellcolor{red!15}93.16 & \cellcolor{red!15}0.47 & \cellcolor{red!15}0.75 & 2.01 \\
            \hline
        \end{tabular}
    }
    \caption{\textbf{Ablation Study on Semantic Enhancement and Semantic-aware Sampling.} We found semantic enhancement improves the semantic correlation and fidelity for the text-to-motion domain. And semantic-aware sampling brings semantic correlation gain with less fidelity and diversity loss on three domains. $\colorbox{red!15}{\rm Indicate best results}$.}
    \label{tab:ablation-sas}
\end{table*}

\begin{figure}[h]
    \centering
    \includegraphics[width=1.0\linewidth]{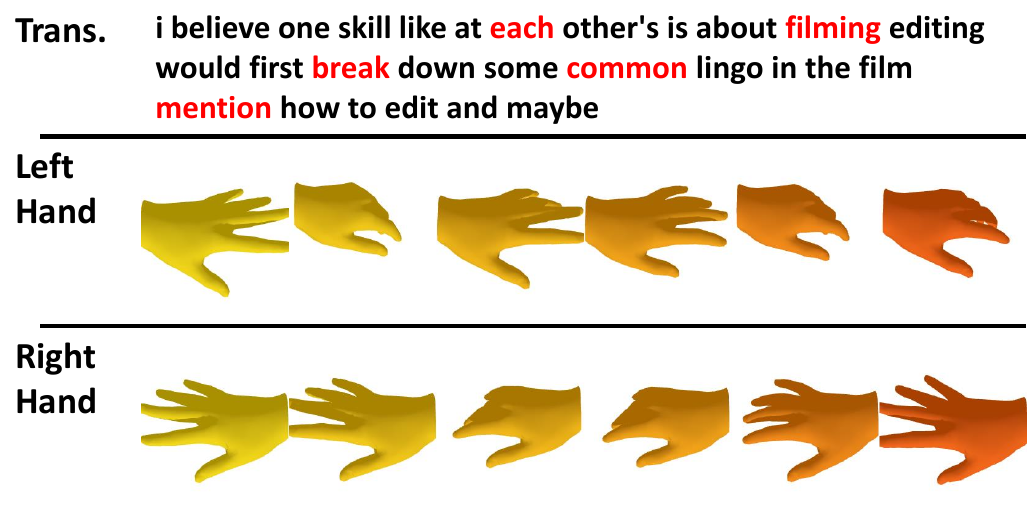}
    \caption{\textbf{Hands Motion Generation.} We evenly visualize 6 poses from a 10-second motion. The transcription is also shown at the top row and keywords corresponding to selected frames are highlighted.}
    \label{fig:hand}
\end{figure}

\subsection{Datasets and Preprocessing}We utilize HumanML3D\cite{guo2022generating} and AIST++\cite{li2021ai}  for text-conditioned and music-conditioned task respectively. For speech-conditioned tasks, we use BEAT\cite{liu2022beat}, a large-scale conversational gesture synthesis dataset. We follow \cite{zhou2023ude} to preprocess the HumanML3D and AIST++ datasets, resulting in motion sequences with identical representation. For BEAT\cite{liu2022beat}, we use the same motion representation for the body sequences, and for left hand and right hand, we use MANO\cite{MANO:SIGGRAPHASIA:2017} representation with 12 PCA components for each hand. Because BEAT has a large amount of data, we only use data corresponding to 4 IDs for training and evaluation(ID: 2, 4, 6, 8). For music and speech data, we use raw acoustic wave sampled as 16KHz.\par

\subsection{Implementation Details}For tokenization, the codebook size of body parts is set to 1024 for all text-, music-, and speech-conditioned generations. The codebook size of hand parts is 512. The embedding dimensions are 512 for all. We train 500 epochs for VQ-VAEs. For condition-to-motion transform, the encoder part is an 8-layer transformer encoder, the base decoder part is a 6-layer transformer decoder, and the head decoder parts are a 2-layer transformer decoder, respectively. We don't train our model on multi-domain data jointly at first. We train it on HumanML3D\cite{guo2022generating} for 200 epochs, then we integrate AIST++\cite{li2021ai} to continue training for 1000 epochs, then we add BEAT\cite{liu2022beat} for another 1000 epochs.\par

\subsection{Evaluation Metrics}
\paragraph{Text-to-Motion}We use Frechet Inception Distance(\textbf{FID}) to measure the distance between generated motion and the real motion distributions, and diversity(\textbf{Div}) to measure the feature distance among all generated motion sequences. We use multimodality (\textbf{MM}) to measure the diversity of motions conditioned on the same textual description. Finally, we measure the motion-text semantic correlation using \textbf{R-Precision}\cite{guo2022generating}. Inspired by \cite{petrovich2023tmr}\cite{lu2023humantomato}, we train a text-motion alignment model for evaluating R-Precision. The detail of this model is described in Appendix.\par
\vspace{-3mm}

\paragraph{Music-to-Motion}We use \textbf{FID}, \textbf{Div} and \textbf{MM} for motion quality, diversity measurement. We use beat alignment score(\textbf{BAS})\cite{li2021ai} to measure the alignment between motion and music beat. In addition, we use \textbf{MDSC}\cite{zhou2023mdsc} to measure the stylistic alignment between generated dance movements and music pieces.\par
\vspace{-4mm}

\paragraph{Speech-to-Motion} Similarly, we use \textbf{FID} for quality measurement, \textbf{Div} and \textbf{MM} for diversity measurement. We observe that personality affects the motion patterns a lot in speech-driven scenarios, it is beneficial if we can measure the ID consistency. Inspired by \cite{zhou2023mdsc}, we proposed a metric to measure the alignment between motion and ID. We describe the detail in Appendix.\par

\subsection{Results}We evaluate our method on three tasks, 1) text-to-motion, 2) music-to-motion, 3) speech-to-motion, and compare our method with several SOTAs\cite{petrovich2022temos, guo2022tm2t, tevet2022human, zhang2023t2m, zhou2023ude, chen2023executing} respectively. We discuss them in detail as follows.

\paragraph{Comparison on Text-to-Motion}For the text-to-motion task, We reproduce the SOTAs following their official implementations using rotation vector as motion representation. We argue that this setting eliminates the effect of different motion representations. The comparison results are reported in Tab. \ref{tab:main-t2m}, which suggests our method has a better capability of mapping the semantic meaning of condition language descriptions to motion sequences. Meanwhile, our method achieves better motion quality and diversity compared with SOTAs. Fig. \ref{fig:result-t2m} shows a few examples for visual comparison.
\vspace{-3mm}

\paragraph{Comparison on Music-to-Motion}We reproduce \cite{li2021ai, siyao2022bailando, tseng2023edge, zhou2023ude} to compare with ours. For every piece of condition music, we generate dance sequences with the same length as input music using our method and SOTAs. We crop the motion sequences to consecutive 5sec segments and evaluate the metrics on these segments. Tab. \ref{tab:main-a2m} shows that our method has stronger ability in generating style-consistent dance movements from music pieces than others while achieving competitive results in fidelity and diversity. Fig. \ref{fig:result-a2m} also provides visual comparison results to facilitate assessment. 
\vspace{-3mm}

\paragraph{Comparison on Speech-to-Motion}We compare ours with \cite{yang2023diffusestylegesture} on BEAT\cite{liu2022beat} using SMPL-X\cite{SMPL-X:2019} representation. It contains torso and left/right-hand motions. We generate long whole-body motion sequences and crop to multiple 5sec segments for evaluation. We show the results in Tab. \ref{tab:main-s2m} and the visual comparison in Fig. \ref{fig:result-s2m}. Specifically, we show in Fig. \ref{fig:hand} that our method can synthesize vivid and smooth hand motions from speech signals. 

\subsection{Ablation Study} For ablations, we validated the effectiveness of the two-stage VQ-VAE and the weight re-initialization technique. Tab. \ref{tab:ablation-vq} shows that applying both techniques boosts the reconstruction quality of VQ-VAE on all three datasets.\par 

We also investigate the effectiveness of two modules proposed in condition-to-motion transform, 1) semantic enhancement(SemE.), and 2) semantic-aware sampling(SaS.). The ablation results are presented in Tab. \ref{tab:ablation-sas}. We found that SemS. consistently improves the semantic correlation and fidelity for the text-to-motion domain, while this tendency is not obvious for music-conditioned and speech-conditioned domains. We will discuss this in detail in the Appendix. The SaS. brings semantic correlation gain with limited fidelity and diversity loss on three domains.

\section{Conclusion}In this paper, we propose a unified framework that supports synthesizing whole-body human motions with up to 3 modalities of control signals. We adopt separate codebooks and use a two-stage VQ-VAE and re-initialization strategy to improve the motion quantization quality. In addition, we utilize an encoder-decoder architecture to transform multimodal conditions into motion tokens. Meanwhile, we propose a semantic enhancement module to advance the semantic relevance and also introduce a semantic-aware sampling technique to further boost the correlation with limited fidelity and diversity loss. Extensive evaluation suggests our method achieves state-of-the-art in all domains.


{
    \small
    \bibliographystyle{ieeenat_fullname}
    \bibliography{main}
}

\clearpage
\appendix

\twocolumn[
\begin{@twocolumnfalse}
\section*{\centering{Supplementary Material}}
\end{@twocolumnfalse}
]

\begin{figure*}
    \centering
    \includegraphics[width=1.0\linewidth]{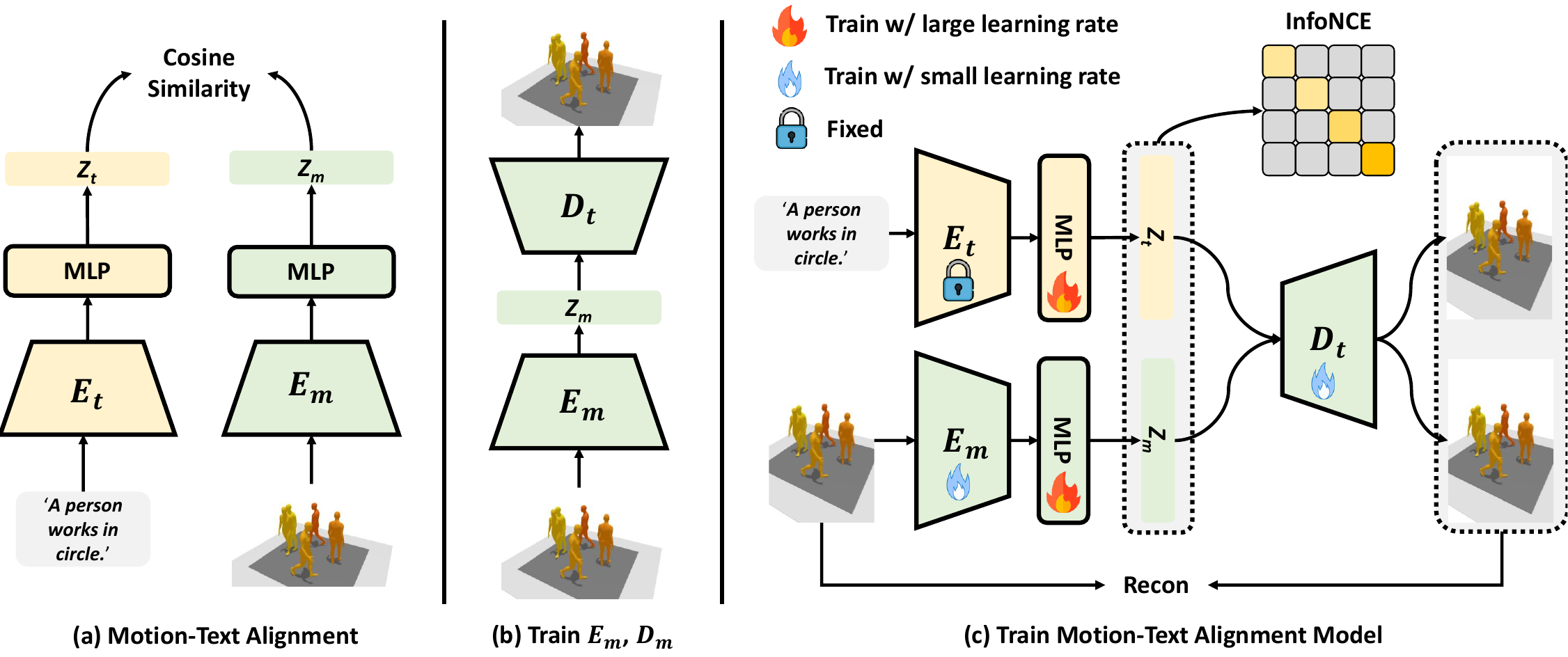}
    \caption{\textbf{Pipeline of Text-Motion Alignment Model.} (a) Measuring the cosine similarity as text-motion alignment. (b) Pretrain motion auto-encoder. (c) Train the text-motion alignment model using reconstruction and infoNCE loss.}
    \label{fig:supp-t2m-metric}
\end{figure*}

\begin{figure*}
    \centering
    \includegraphics[width=1.0\linewidth]{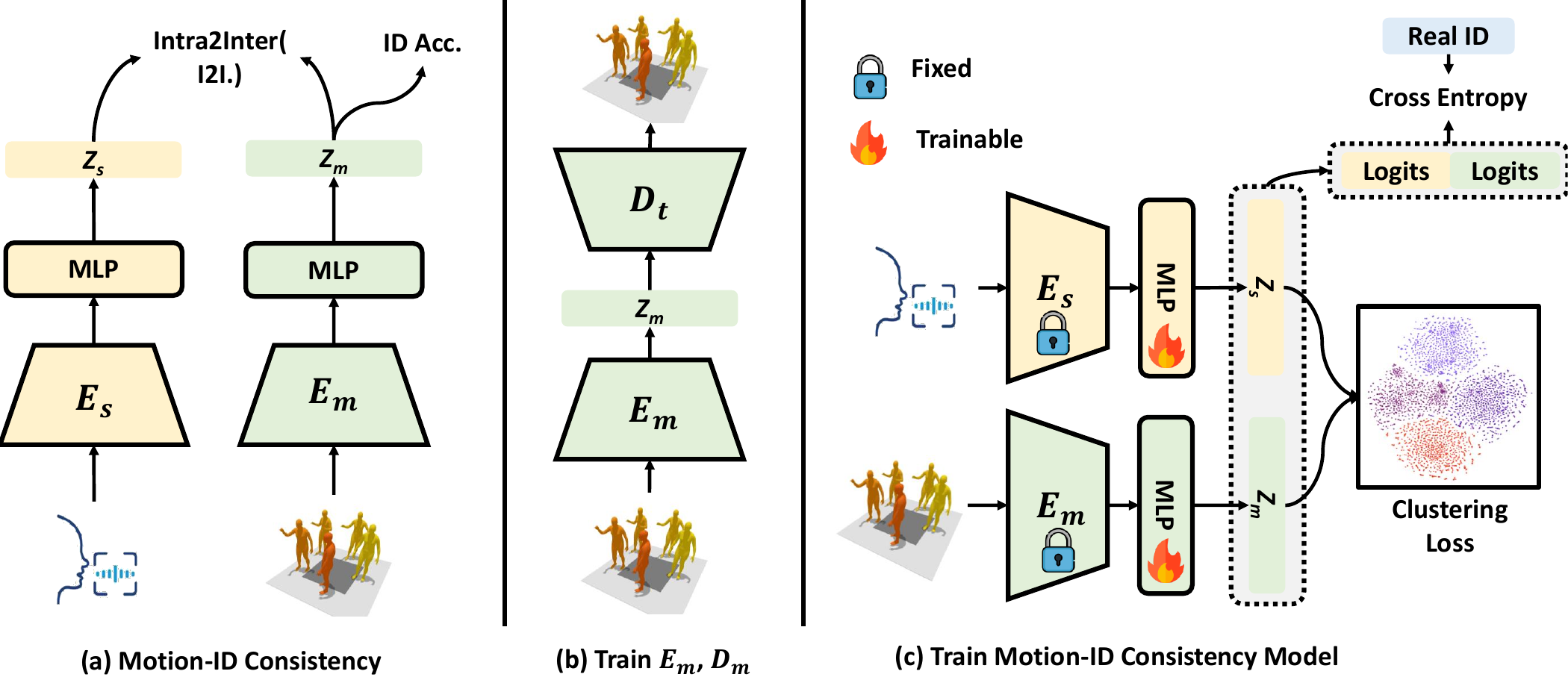}
    \caption{\textbf{Pipeline of Speech-to-Motion ID Consistency Model.} (a) Measuring the Intra2Inter(I2I.) and ID prediction accuracy(ID Acc.) as metrics. (b) Pretrain motion auto-encoder. (c) Train the speech-to-motion ID consistency model using clustering loss and cross entropy loss.}
    \label{fig:supp-s2m-metric}
\end{figure*}

\begin{figure*}
    \centering
    \includegraphics[width=1.0\linewidth]{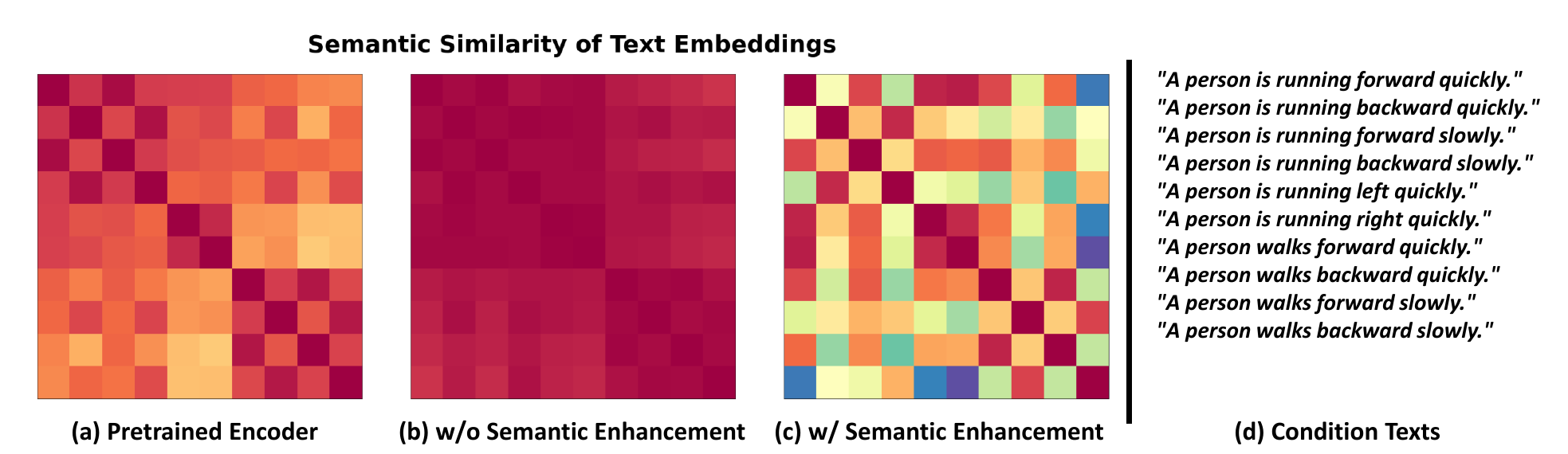}
    \caption{\textbf{Semantic Enhancement on Text Embedding.} We show the pairwise similarity of embeddings obtained by: (a) pretrained text encoder, (b) multimodal condition encoder trained w/o semantic enhancement, (c) multimodal condition encoder trained w/ semantic enhancement. (d) is the condition text descriptions.}
    \label{fig:supp-text-semantic-enhancement}
\end{figure*}

\begin{figure}
    \centering
    \includegraphics[width=1.0\linewidth]{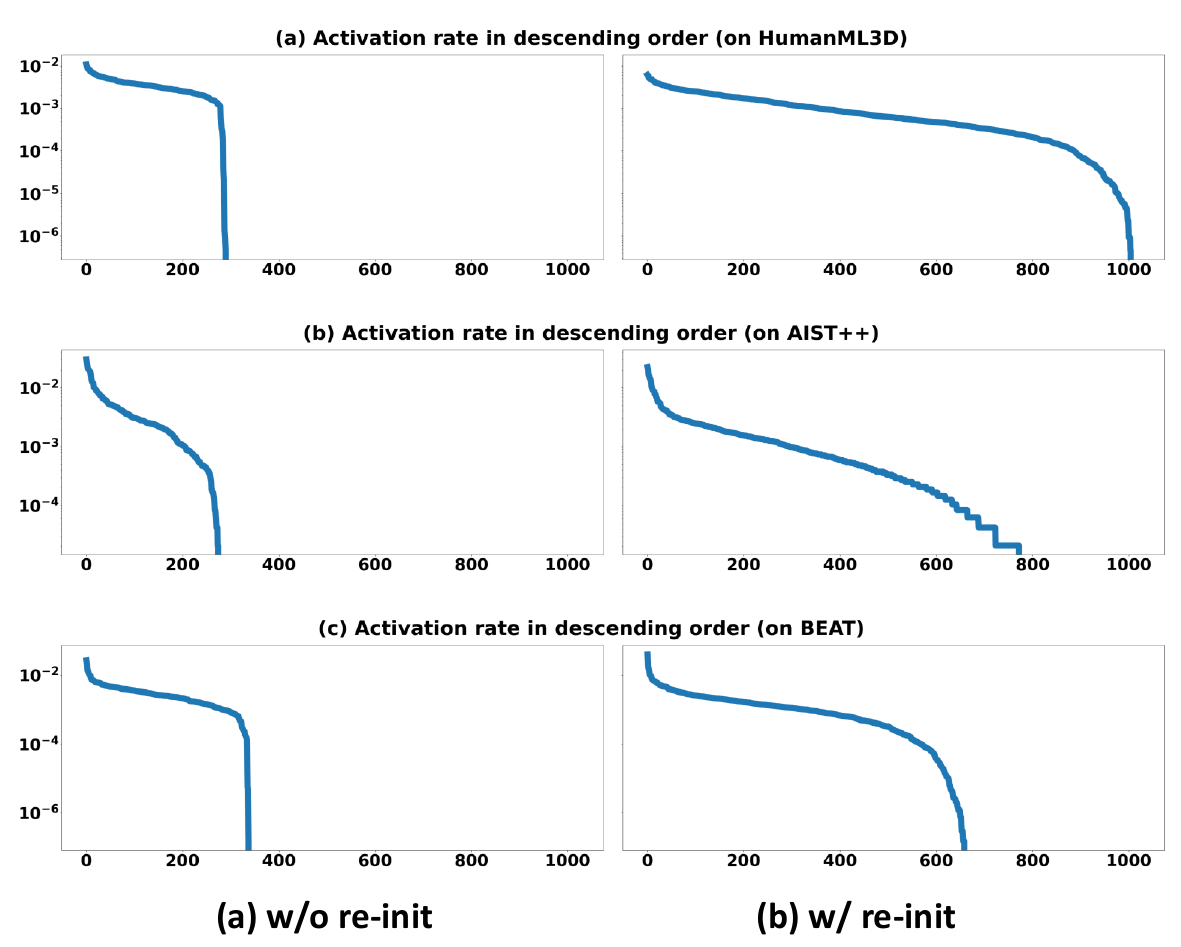}
    \caption{\textbf{Comparison of Activation Rate.} We investigate the effectiveness of weights re-initialization. We sort the activation rate in descending order. We use logarithm coordinates for y-axis.}
    \label{fig:supp-activation-rate}
\end{figure}

\begin{figure*}
    \centering
    \includegraphics[width=0.8\linewidth]{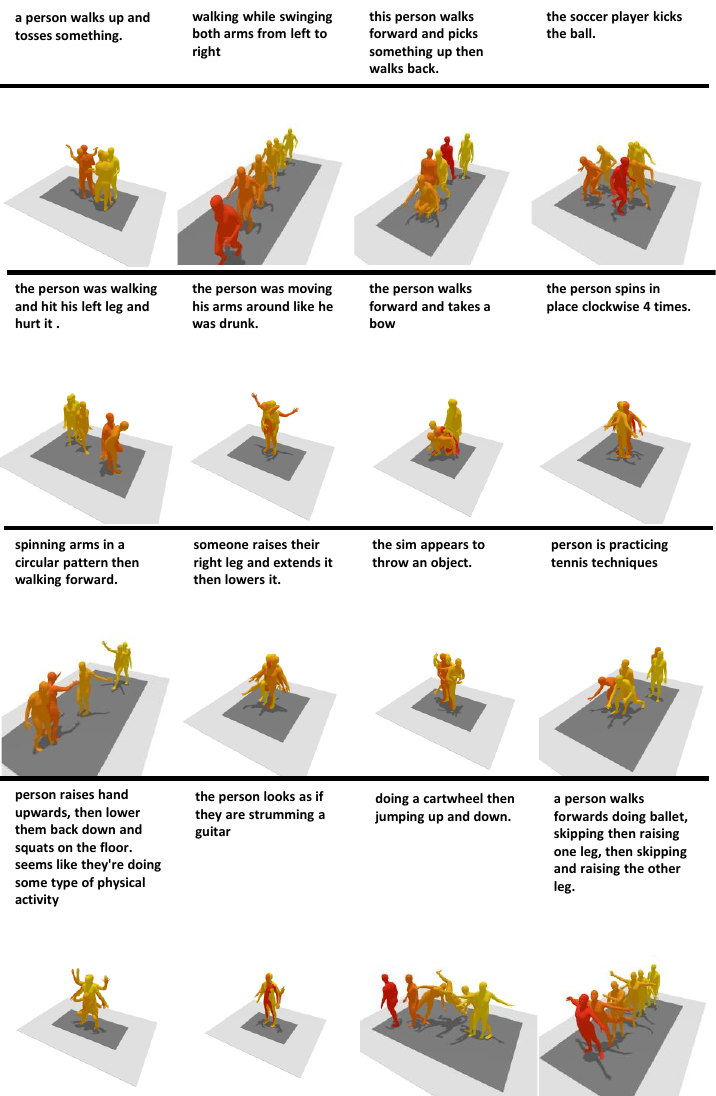}
    \caption{\textbf{More Results on Text-to-Motion Generation.}}
    \label{fig:supp-more-t2m}
\end{figure*}

\begin{figure*}
    \centering
    \includegraphics[width=0.85\linewidth]{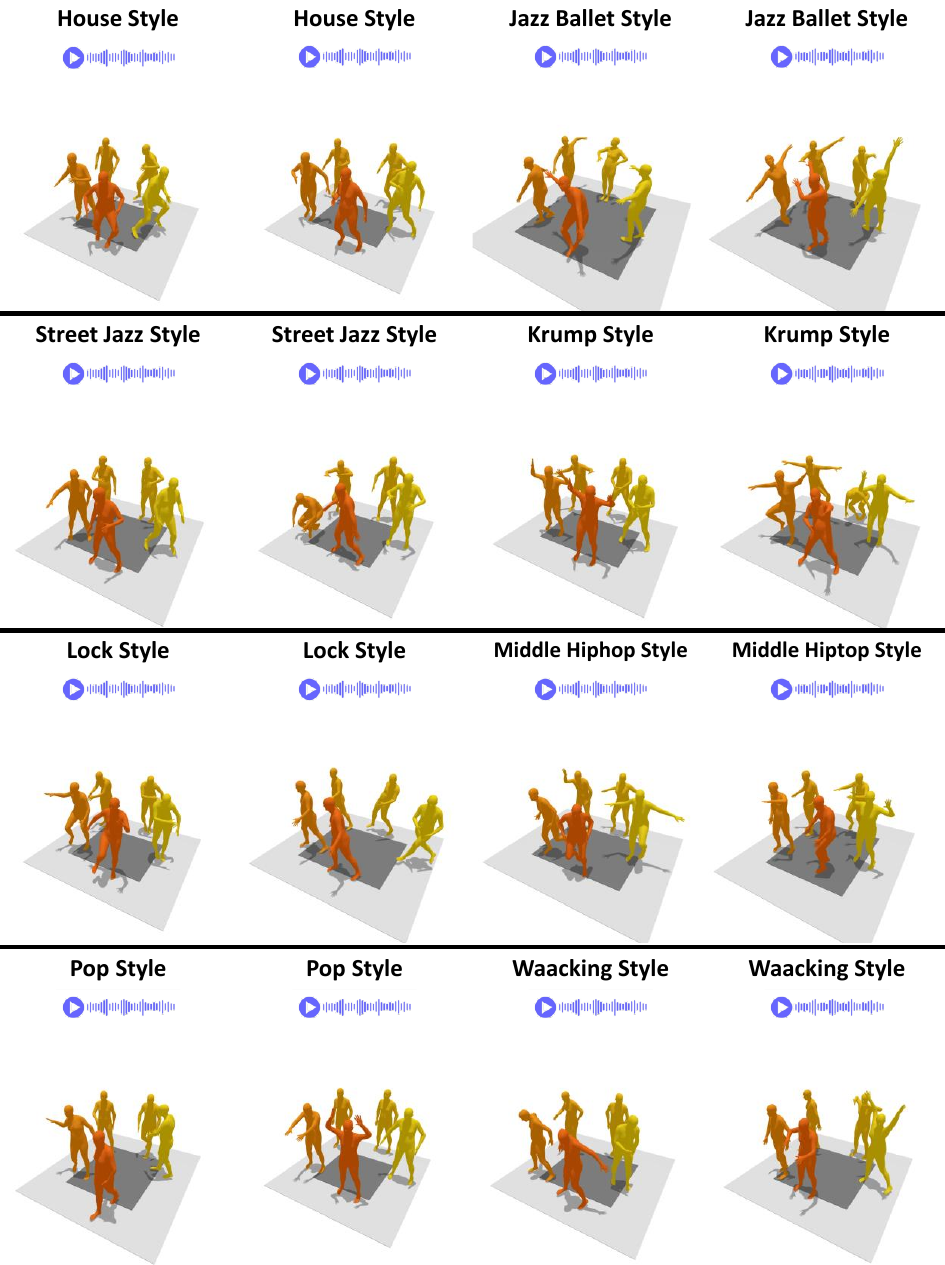}
    \caption{\textbf{More Results on Music-to-Motion Generation.}}
    \label{fig:supp-more-a2m}
\end{figure*}

\begin{figure*}
    \centering
    \includegraphics[width=0.9\linewidth]{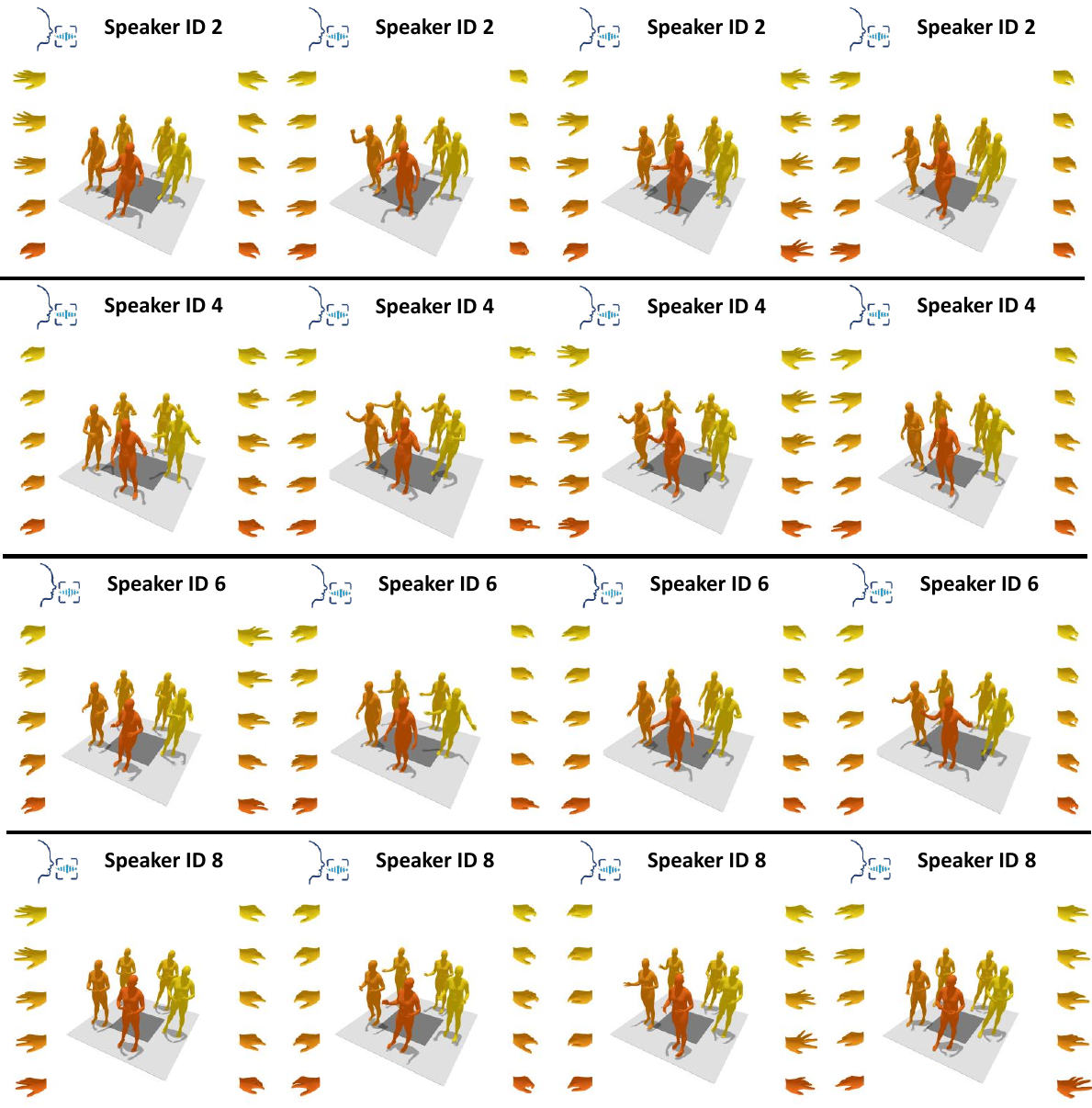}
    \caption{\textbf{More Results on Speech-to-Motion Generation.}}
    \label{fig:supp-more-s2m}
\end{figure*}

\begin{figure*}
    \centering
    \includegraphics[width=0.9\linewidth]{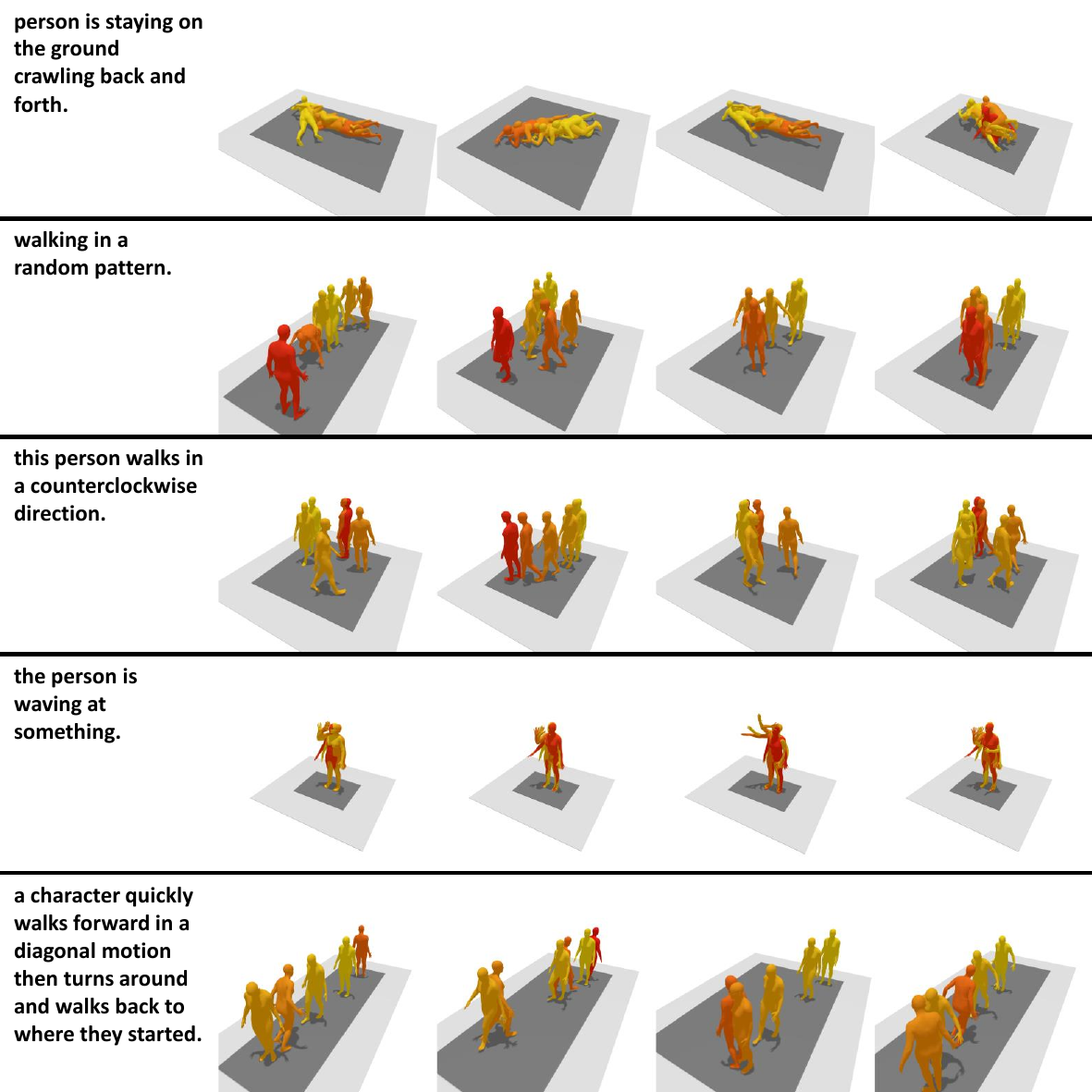}
    \caption{\textbf{More Results on Diverse Text-to-Motion Generation.}}
    \label{fig:supp-div-t2m}
\end{figure*}

\begin{figure*}
    \centering
    \includegraphics[width=0.9\linewidth]{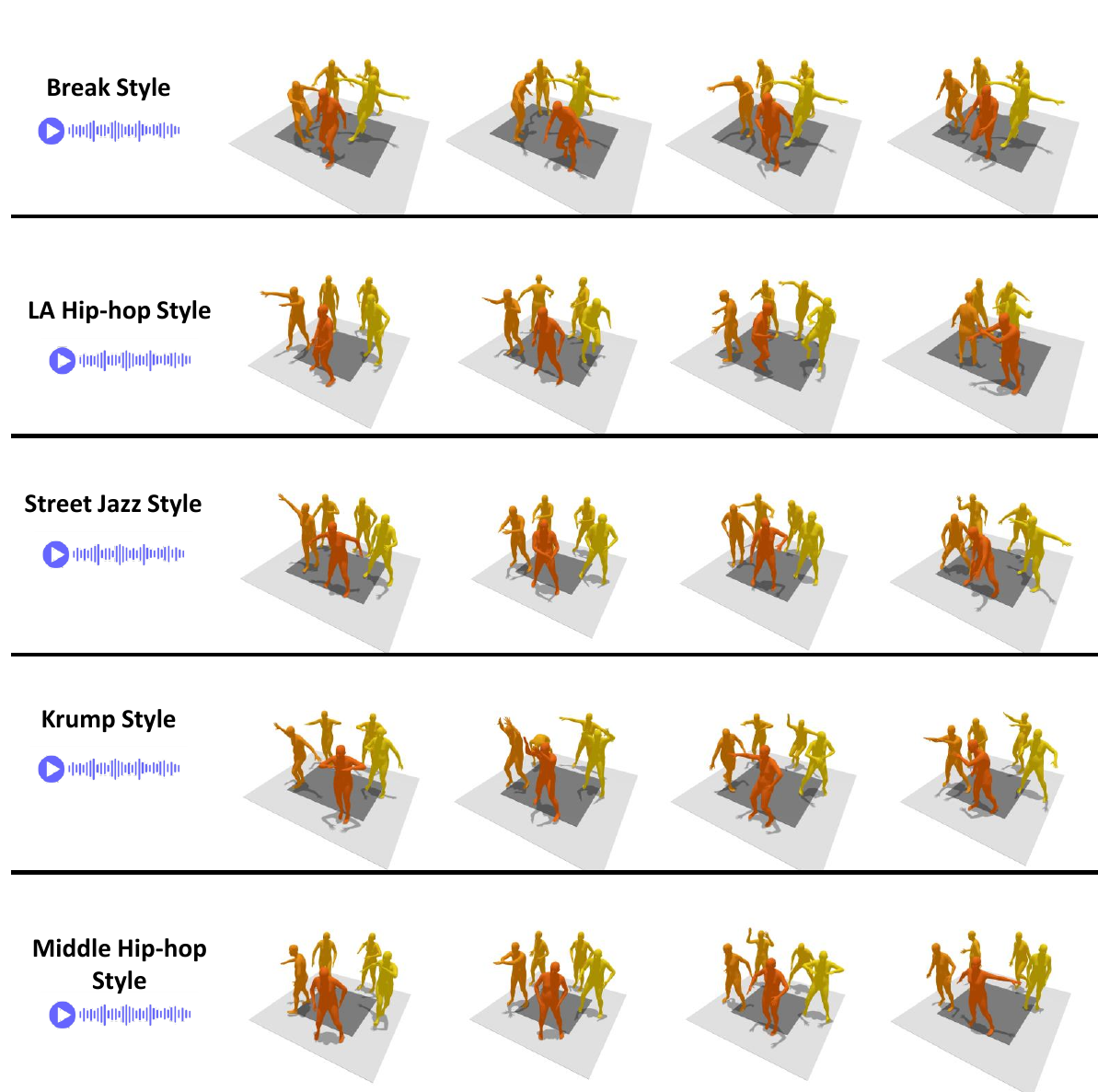}
    \caption{\textbf{More Results on Diverse Music-to-Motion Generation.}}
    \label{fig:supp-div-a2m}
\end{figure*}

\begin{figure*}
    \centering
    \includegraphics[width=1.0\linewidth]{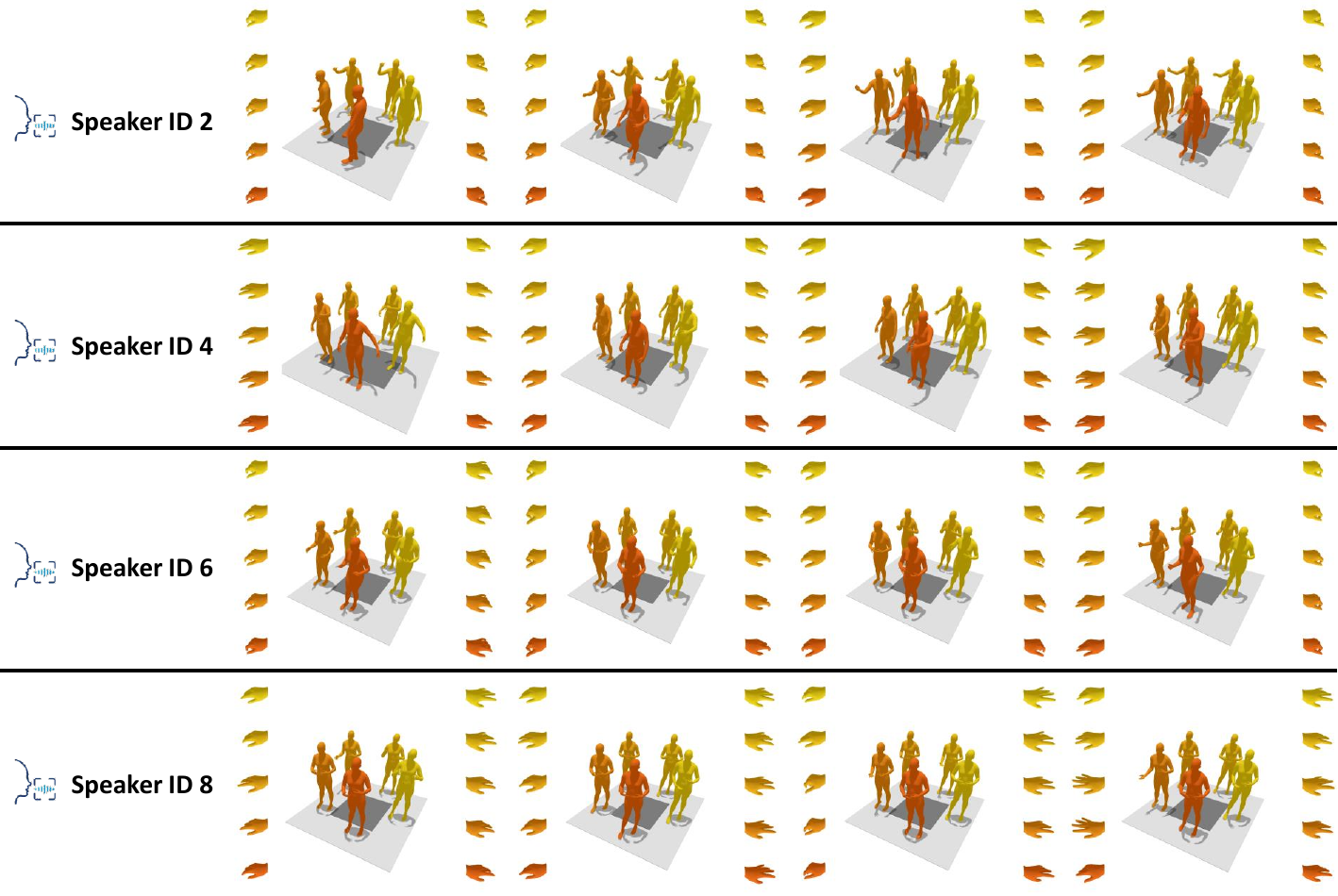}
    \caption{\textbf{More Results on Diverse Speech-to-Motion Generation.}}
    \label{fig:supp-div-s2m}
\end{figure*}

\section{Text-Motion Alignment Model}Fig. \ref{fig:supp-t2m-metric} shows our approach in evaluating motion-to-text consistency. We encode the motion sequence and text description to embeddings $z_m$ and $z_t$, respectively. And measure the cosine similarity as the degree of alignment between motion and text. We first train a motion auto-encoder and use its encoder to encode the motion sequence. We adopt the model design proposed in \cite{petrovich2021action}. For the text encoder, we use the pretraind CLIP\cite{radford2021learning} text encoder to encode the text description. As discussed previously(Section $\color{red}{3.2}$), the semantic discrepancy between motion and text exists. We show how to train the text-motion alignment model in Fig. \ref{fig:supp-t2m-metric} (c). We introduce two adapter layers, they transform the embeddings obtained by motion and text encoder $z_m$ and $z_t$ to the same dimension. To align them in semantically meaningful latent space, we regularize them through reconstructing motion from aligned embeddings. During the training, we initialize the encoders and decoders with their pretrained weights. We fix the CLIP text encoder, and we tune the pretrained motion encoder and decoder with a small learning rate. We train the adapter layers with a relatively large learning rate. The alignment model is trained by optimizing the following objective:
\vspace{-3mm}

\begin{equation}
    \mathcal{L} = \mathcal{L}_{recon} + \mathcal{L}_{infoNCE}
    \label{eq:supp-obj}
\end{equation}
\vspace{-3mm}

Where\par 
\vspace{-3mm}
\begin{equation}
    \mathcal{L}_{recon}=\|x - \tilde{x}\|_2
    \label{eq:supp-recon}
\end{equation}
\vspace{-3mm}
is the reconstruction term, and\par
\vspace{-3mm}

\begin{equation}
    \mathcal{L}_{infoNCE}=-\frac{1}{N}\Sigma_{i=1}^K(\log{\frac{exp(\langle z_m^i, z_t^i \rangle / \tau)}{exp(\Sigma_{j=1}^K)(\langle z_m^i, z_t^j \rangle / \tau)}})
    \label{eq:supp-infonce}
\end{equation}

is the infoNCE loss which helps align paired text-motion close while pushing unpaired apart.\par

We compare different models and training designs as motion-text alignment models in Tab. \ref{tab:supp-t2m-ablation}. It is noticeable that tuning pretrained $E_m$ and $D_m$ with a small learning rate brings remarkable performance improvement, which proves the effectiveness of our design.

\section{Speech-to-Motion ID Consistency Evaluation Model}We argue that the characters of speakers affect the motion patterns severely in terms of the speech-to-motion domain. We propose a novel method to evaluate ID consistency. Inspired by \cite{zhou2023mdsc}, we formulate it as a clustering problem. Specifically, if motions could be encoded to latent representation embeddings, embeddings corresponding to the motions performed by the same person should be close to each other in latent space, while those corresponding to different persons should be away from each other. This is a typical clustering hypothesis. Fig. \ref{fig:supp-s2m-metric} shows the proposed model. We propose two metrics to assess ID consistency. 1) I2I: measures the ratio between intra-cluster to inter-cluster distance. Given a target motion, its representation is encoded by pretrained motion encoder as $z_m$. We calculate the intra-cluster distance as: $d_{intra} = \|z_m - \hat{z}_m^i\|_2$, where $\hat{z}_m^i$ is the learned cluster center of ID $i$, and the inter-cluster distance is calculated as: $d_{inter} = \frac{1}{N-1}\Sigma_{j=1, j \neq i}^N(\|z_m^i - \hat{z}_m^j\|_2)$. Smaller I2I indicates higher ID consistency. 2) Acc: measure the ID prediction accuracy. We first train a motion auto-encoder, then we fix the encoder and a pretrained speech encoder\cite{hsu2021hubert} as encoder priors. We use two adapter layers to align the speech and motion embeddings in latent space. These adapter layers are only trainable modules here, and we train them using clustering loss and cross-entropy loss(Fig. \ref{fig:supp-s2m-metric} (c)).

\section{More Results of Ablation Study}

\begin{table}[]
    \centering
    \resizebox{1.0\linewidth}{!}{
        \begin{tabular}{cc|ccccc}
            \hline
            \multicolumn{2}{c|}{Variants} & \multirow{2}{*}{R Top-1 $\uparrow$} & \multirow{2}{*}{R Top-2 $\uparrow$} & \multirow{2}{*}{R Top-3 $\uparrow$} & \multirow{2}{*}{R Top-4 $\uparrow$} & \multirow{2}{*}{R Top-5 $\uparrow$} \\
            \cline{1-2}
            tune $E_m$ & tune $D_m$ & & & & & \\
            \hline
            \XSolidBrush & \XSolidBrush & 19.85 & 29.18 & 35.25 & 40.58 & 45.18 \\
            \XSolidBrush & \checkmark & 50.96 & 66.13 & 73.67 & 78.72 & 82.40 \\
            \checkmark & \checkmark & \cellcolor{red!15}63.60 & \cellcolor{red!15}79.23 & \cellcolor{red!15}85.02 & \cellcolor{red!15}88.14 & \cellcolor{red!15}90.44 \\
            \hline
        \end{tabular}
    }
    \caption{\textbf{Comparison on Text-Motion Alignment.} We compare different models and training designs on the text-motion alignment model. We found tuning $E_m$ and $D_m$ brings significant performance gain. $\colorbox{red!15}{\rm Indicate best results}$.}
    \label{tab:supp-t2m-ablation}
\end{table}

\paragraph{Weights Re-Initialization}We compare the activation rate of codebook tokens with and without using the weights re-initialization strategy on three datasets. Fig. \ref{fig:supp-activation-rate} presents the comparison results. We encode the test sets and count the activation rate. We sort the results according to the activation rate in descending order and display them in a histogram-like style. It is clearly observed that the weights re-initialization strategy increases the code activation rate remarkably. 

\paragraph{Semantic Enhancement}We investigate the effectiveness of the proposed semantic enhancement module. It is discussed previously that semantic enhancement is particularly effective in the text-motion domain, while large-scale pre-trained models like CLIP\cite{radford2021learning} are not able to capture the semantics accurately in some cases. For instance, Fig. \ref{fig:supp-text-semantic-enhancement} (a) shows the semantic similarity between embeddings encoded by CLIP. The associated textual descriptions are shown in Fig. \ref{fig:supp-text-semantic-enhancement} (d). As we can see, CLIP text encoder fails to understand the key semantics. For example, `\textit{a person walks forward quickly}' and `\textit{a person walks backward quickly}' have very similar expressions, but their semantics vary a lot. But the similarity shows that `\textit{a person walks forward quickly}' is closer to `\textit{a person walks backward quickly}' than `\textit{a person is running forward quickly}'. We believe that this is because CLIP is trained on a pair of static images and text, and the dynamic semantics in the text is not well learned as a result. We show in Fig. \ref{fig:supp-text-semantic-enhancement} (b), (c) that our proposed semantic enhancement loss helps to learn better semantics from textual description. For instance, it helps distinguish the semantics from `\textit{forward}' and `\textit{backward}' accurately. 

\section{More Results on Motion Generation}We present more motion generation results in Fig. \ref{fig:supp-more-t2m}, \ref{fig:supp-more-a2m}, \ref{fig:supp-more-s2m}. For the text-to-motion domain, we show 16 motion sequences synthesized by different textual descriptions. For the music-to-motion domain, we select 8 genres and show two samples per genre. Each sample is a 4-second dance segment. For the speech-to-motion domain, we show samples of the generation of 4 speakers. We display the whole-body motions in the middle, and we enlarge the displays of left/right-hand movements at the side for convenience of visualization. 

\section{More Results of Diverse Generation}We show more results to demonstrate our method's capability in synthesizing diverse motion with high condition consistency in Fig. \ref{fig:supp-div-t2m}, \ref{fig:supp-div-a2m}, \ref{fig:supp-div-s2m}. We synthesize 4 motion sequences from each condition signal to demonstrate the diversity of generation. We show diverse torso movements for text-to-motion and music-to-motion scenarios and present both torch and hand movements for speech-to-motion scenarios. The results suggest that our method is able to synthesize not only diverse torch movements but also diverse and smooth hand movements.

\section{Limitation and Future Work}1) Due to the limitation of dataset\cite{guo2022generating, li2021ai}, our method is currently not available for generating hand movements for text- and music-driven scenarios. Collecting datasets with richer annotation(e.g. \cite{lin2023motionx}), as well as enabling the text- and music-conditioned hand motion synthesis is one of our prior future works. 2) Our method supports synthesizing a single person's motion at this stage. Generating interactive actions conditioned on multimodal signals is of great value. This includes interacting with other persons or with the environment. Our future work also focuses on it. 3) Appropriate evaluation metrics are beneficial to further improvements. Our future work also lies in investigating better metrics.

\end{document}